\documentclass[letterpaper, 10 pt, conference]{ieeeconf}  %

\IEEEoverridecommandlockouts                              %

\usepackage{times}
\usepackage{graphicx}
\usepackage{epsfig}
\usepackage{epstopdf}
\usepackage{amsmath} %
\usepackage{amssymb}  %
\usepackage{amsfonts}  %
\usepackage{mathtools}
\usepackage[mathscr]{euscript}
\usepackage{subfig}
\usepackage[nolist,nohyperlinks]{acronym}
\usepackage[boxed]{algorithm}
\usepackage{algpseudocode}%
\usepackage[super]{nth}
\usepackage[overload]{textcase}
\usepackage{graphics} %
\usepackage{amsmath} %
\usepackage{mathtools}
\usepackage{amssymb}  %
\usepackage{multirow}
\usepackage{siunitx}
\usepackage{xfrac}
\usepackage[protrusion=true,
             expansion=true,
             tracking=true,
             final,
             stretch=25,factor=1250]{microtype}
\title{\LARGE \bf
\acs{sedar}\;\textendash\;Semantic Detection and Ranging: \\ Humans can localise without LiDAR, can robots?\vspace{-0.5cm}
}

\author{
Oscar Mendez\\
University of Surrey\\
Guildford GU2\\
{\tt\small o.mendez@surrey.ac.uk}
\and
Simon Hadfield\\
University of Surrey\\
Guildford GU2\\
{\tt\small s.hadfield@surrey.ac.uk}
\and
Nicolas Pugeault\\
University of Exeter\\
Exeter EX4\\
{\tt\small n.pugeault@exeter.ac.uk}
\and%
Richard Bowden\\
University of Surrey\\
Guildford GU2\\
{\tt\small r.bowden@surrey.ac.uk}
\thanks{The Titan X used for this research was donated by the NVIDIA Corporation.}
}

\DeclareMathOperator*{\mymax}{max}
\DeclareMathOperator*{\myargmax}{arg\,max}
\DeclareMathOperator*{\mymin}{min}
\DeclareMathOperator*{\myargmin}{arg\,min}

\DeclareSymbolFont{bbold}{U}{bbold}{m}{n}
\DeclareSymbolFontAlphabet{\mathbbold}{bbold}

\begin{document}
\begin{acronym}[RANSAC] 
\acro{nbv}[NBV]{Next-Best View}
\acro{nbs}[NBS]{Next-Best Stereo}

\acro{bmvc}[BMVC]{British Machine Vision Conference}
\acro{iccv}[ICCV]{International Conference on Computer Vision}

\acro{agast}[AGAST]{Adaptive and Generic Accelerated Segment Test}
\acro{fast}[FAST]{Features from Accelerated Segment Test}

\acro{ate}[A]{Absolute Trajectory Error}
\acro{anne}[ANNE]{Average Nearest-Neighbour Error}
\acro{rpe}[RPE]{Relative Pose Error}
\acro{rmse}[RMSE]{Root Mean Square Error}

\acro{dof}[DoF]{Degrees of Freedom}
\acro{cs}[\ensuremath{\acs{confspace}}]{Configuration Space}
\acro{se2}[SE(2)]{Special Euclidean Space}
\acro{so2}[SO(2)]{Special Orthogonal Space}
\acro{se3}[SE(3)]{Special Euclidean Space}
\acro{so3}[SO(3)]{Special Orthogonal Space}

\acro{cnn}[CNN]{Convolutional Neural Network}
\acro{lut}[LUT]{Lookup Table}
\acro{ls}[LS]{Least-Squares}
\acro{kdt}[k-d tree]{k-dimensional Tree}
\acro{if}[IF]{Information Filter}
\acro{lls}[L-LS]{Linear Least-Squares}
\acro{ills}[ILLS]{Iterative Linear-Least-Squares}
\acro{mvs}[MVS]{Multi-View Stereo}
\acro{mocap}[MoCap]{Motion Capture}
\acro{pid}[PID]{Proportional Integral Controller}
\acro{vp}[VP]{Vanishing Point}
\acro{svd}[SVD]{Single Value Decomposition}
\acro{ba}[BA]{Bundle Adjustment}
\acro{sf}[SF]{Sensor-Fusion}
\acro{sfm}[SfM]{Structure from Motion}
\acro{csfm}[CSfM]{Collaborative Structure from Motion}
\acro{vo}[VO]{Visual Odometry}

\acro{klf}[KF]{Kalman Filter}
\acro{ekf}[EKF]{Extended Kalman Filter}
\acro{ukf}[UKF]{Unscented Kalman Filter}

  \acro{icp}[ICP]{Iterative Closest Point}
  \acro{lsd}[LSD]{Line Segment Detector}
  \acro{pnp}[PnP]{Perspective N-Points}
  \acro{ransac}[RANSAC]{RANdom Sample And Consensus}
  \acro{rsc}[RANSAC]{RANdom SAmple and Consensus}
  \acro{svo}[SVO]{Semi-Direct Visual Odometry}
  \acro{vsfm}[VSFM]{Visual Structure from Motion}
  
  \acro{slam}[SLAM]{Simultaneous Localization and Mapping}
  \acro{vslam}[VSLAM]{Vision-Based Simultaneous Localization and Mapping}
  
  \acro{etam}[ETAM]{ETHZASL-PTAM}
  \acro{ptam}[PTAM]{Parallel Tracking and Mapping}
  \acro{dtam}[DTAM]{Dense Tracking and Mapping}

  \acro{pf}[PF]{Particle Filter}
  \acro{rbpf}[RBPF]{Rao-Blackwellised Particle Filter}
  \acro{smc}[SMC]{Sequential Monte-Carlo}
  \acro{mcl}[MCL]{Monte-Carlo Localisation}
  \acro{amcl}[AMCL]{Adaptive Monte Carlo Localisation}
  \acro{vmcl}[VMCL]{Vision-Based Monte-Carlo Localisation}
  \acro{rmcl}[RMCL]{Range-Based Monte-Carlo Localisation}
  
  \acro{prm}[PRM]{Probabilistic Road Map}
  \acro{rrt}[RRT]{Rapidly-exploring Random Tree}
  \acro{rrt*}[RRT*]{Rapidly-exploring Random Tree}

\acro{ompl}[OMPL]{Open Motion Planning Libraries}
\acro{pcl}[PCL]{Point Cloud Library}
\acro{ros}[ROS]{Robot Operating System}

\acro{gps}[GPS]{Global Positioning System}
\acro{imu}[IMU]{Inertial Measurement Unit}
\acro{lidar}[LiDAR]{Light Detection And Ranging}
\acro{rgb}[RGB]{Red, Green and Blue}
\acro{rgbd}[RGB-D]{\acs{rgb} and Depth}
\acro{rgbds}[RGB-DS]{\acs{rgb}, Depth and Semantic}
\acro{sedar}[SeDAR]{Semantic Detection and Ranging}
\acro{smcl}[SeMCL]{Semantic Monte-Carlo Localisation}
\acro{sonar}[SONAR]{SOund Navigation And Ranging}

\acro{ada}[ADA]{ARDrone Autonomy}
\acro{ai}[AI]{Artificial Intelligence}
\acro{ar}[AR]{Augmented Reality}
\acro{ard}[ARD]{AR Drone}
\acro{mct}[MTC]{MATLAB Calibration Toolbox}
\acro{tma}[TMA]{TUM\_ARDrone}
\acro{tum}[TUM]{Technical University of Munich}
\acro{uav}[UAV]{Unmanned Aerial Vehicle}
\acro{sdk}[SDK]{Software Development Kit}
\acro{fps}[FPS]{Frames Per Second}
\end{acronym}

%

\mathchardef\mhyphen="2D

\def\eg{\emph{e.g}.\ } \def\Eg{\emph{E.g}.\ }
\def\ie{\emph{i.e}.\ } \def\Ie{\emph{I.e}.\ }
\def\cf{\emph{c.f}.\ } \def\Cf{\emph{C.f}.\ }
\def\etc{\emph{etc}.\ } \def\vs{\emph{vs}.\ }
\def\wrt{w.r.t.\ } \def\dof{d.o.f.\ }
\def\etal{\emph{et al}.\ }


\DeclareRobustCommand{\mysmallmath}[1]{\scriptscriptstyle{#1}}
\newcommand{\mysmallscript}[1]{\textnormal{\tiny \textsc{#1}}}
\newlength{\depthofsumsign}
\setlength{\depthofsumsign}{\depthof{$\sum$}}
\newlength{\totalheightofsumsign}
\newlength{\heightanddepthofargument}

\newcommand{\nsum}[1][1.4]{
    \mathop{%
        \raisebox
            {-#1\depthofsumsign+1\depthofsumsign}
            {\scalebox
                {#1}
                {$\displaystyle\sum$}%
            }
    }
}

\newcommand{\distas}[0]{\sim}

\newcommand{\e}[0]{e}
\newcommand{\myexp}[1]{\ensuremath{\sca{}{}{\e}{}{^{#1}}}}
\newcommand{\minus}{\scalebox{0.5}[0.75]{\ensuremath{-}}}
\newcommand{\inverse}{\mysmallmath{{\minus1}}}
\newcommand{\slfrac}[2]{\left.#1\middle/#2\right.}

\newcommand{\brackets}[3]{\ensuremath{\left#1 #2 \right#3}}
\newcommand{\mypar}[1]{\brackets{(}{#1}{)}}
\newcommand{\mybra}[1]{\brackets{\{}{#1}{\}}}
\newcommand{\mymag}[1]{\brackets{\|}{#1}{\|}}
\newcommand{\mynrm}[1]{\brackets{|}{#1}{|}}
\newcommand{\mysqb}[1]{\brackets{[}{#1}{]}}

\newcommand{\idx}[2][]{\ifthenelse{\equal{#1}{}}{#2}{#2{#1}}}
\newcommand{\mytime}[1][]{\idx[#1]{t}}
\newcommand{\myTime}[1][]{\idx[#1]{T}}
\newcommand{\iindex}[1][]{\idx[#1]{i}}
\newcommand{\Iindex}[1][]{\idx[#1]{I}}
\newcommand{\jindex}[1][]{\idx[#1]{j}}
\newcommand{\Jindex}[1][]{\idx[#1]{J}}
\newcommand{\kindex}[1][]{\idx[#1]{k}}
\newcommand{\Kindex}[1][]{\idx[#1]{K}}

\newcommand{\mysym}[5]{\ensuremath{\prescript{#1}{#2\hspace{+0.25mm}}{#3}_{#4}^{#5}}}

\newcommand{\mysca}[1]{\ensuremath{\lowercase{#1}}}
\newcommand{\myvtr}[1]{\ensuremath{\mathbf{\lowercase{#1}}}}
\newcommand{\mywht}[1]{\ensuremath{{\lowercase{#1}}}}
\newcommand{\myset}[1]{\ensuremath{\mathbb{\MakeUppercase{#1}}}}
\newcommand{\mymat}[1]{\ensuremath{\mathbf{\uppercase{#1}}}}
\newcommand{\myfunc}[1]{\ensuremath{\boldsymbol{\lowercase{#1}}}}

\newcommand{\sca}[5]{\mysym{#1}{#2}{\mysca{#3}}{#4}{#5}}
\newcommand{\vtr}[5]{\mysym{#1}{#2}{\myvtr{#3}}{#4}{#5}}
\newcommand{\whtt}[5]{\mysym{#1}{#2}{\mywht{#3}}{#4}{#5}}
\newcommand{\set}[5]{\mysym{#1}{#2}{\myset{#3}}{#4}{#5}}
\newcommand{\mat}[5]{\mysym{#1}{#2}{\mymat{#3}}{#4}{#5}}
\newcommand{\fnc}[5]{\mysym{#1}{#2}{\myfunc{#3}}{#4}{#5}}
\newcommand{\func}[6]{\fnc{#1}{#2}{#3}{#4}{#5}\mypar{#6}}
\newcommand{\easyfunc}[2]{\func{}{}{#1}{}{}{#2}}

\newcommand{\tra}[4]{\mat{#1}{#2}{T}{#3}{#4}}
\newcommand{\img}[4]{\mat{#1}{#2}{I}{#3}{#4}}
\newcommand{\flo}[4]{\mat{#1}{#2}{F}{#3}{#4}}
\newcommand{\prj}[4]{\mat{#1}{#2}{P}{#3}{#4}}

\newcommand{\myunion}[2]{\ensuremath{\bigcup_{\mysmallmath{\begin{substack}{ #1 }\end{substack}}} #2}}
\newcommand{\mytuple}[1]{\ensuremath{\left\langle #1 \right\rangle}}
\newcommand{\myprod}[4]{\ensuremath{\prod_{#1=#2}^{#3}{#4}}}
\newcommand{\myprob}[2]{\ensuremath{\Pr_{#2}\mypar{#1}}}
\newcommand{\mygaussian}[2]{\ensuremath{\mathcal{N}\left(#1,#2\right)}}			

\newcommand{\valmin}[2]{\mymin_{#1}\hspace{1mm}#2}
\newcommand{\argmin}[2]{\myargmin_{#1}\hspace{1mm}#2}
\newcommand{\valmax}[2]{\mymax_{#1}\hspace{1mm}#2}
\newcommand{\argmax}[2]{\myargmax_{#1}\hspace{1mm}#2}

\newcommand{\thr}[1]{\sca{}{}{\tau}{#1}{}}
\newcommand{\cst}[2]{\sca{}{}{#1}{#2}{}}

\newcommand{\psn}[0]{x}
\DeclareRobustCommand{\posn}[2]{\vtr{}{}{\psn}{#1}{#2}}
\DeclareRobustCommand{\posnSet}[2]{\set{}{}{\psn}{#1}{#2}}

\newcommand{\pos}[0]{\dot{\psn}}
\DeclareRobustCommand{\pose}[2]{\vtr{}{}{\pos}{#1}{#2}}
\DeclareRobustCommand{\poseSet}[2]{\set{}{}{\pos}{#1}{#2}}

\DeclareRobustCommand{\poseWorld}[2]{\vtr{}{w}{\pos}{#1}{#2}} 

\newcommand{\pgt}[0]{g}
\DeclareRobustCommand{\posngt}[2]{\vtr{}{}{\pgt}{#1}{#2}}
\DeclareRobustCommand{\posngtSet}[2]{\set{}{}{\pgt}{#1}{#2}}

\newcommand{\px}[0]{m}
\newcommand{\pxs}[4]{\vtr{#1}{#2}{\px}{#3}{#4}}
\newcommand{\pxsSet}[4]{\set{#1}{#2}{\px}{#3}{#4}}

\newcommand{\pt}[0]{\dot{\px}}
\DeclareRobustCommand{\pts}[4]{\vtr{#1}{#2}{\pt}{#3}{#4}}
\DeclareRobustCommand{\ptsSet}[4]{\set{#1}{#2}{\pt}{#3}{#4}}

\DeclareRobustCommand{\ptsVox}[3]{\pts{#1}{#2}{\vox{}{}{}{}}{#3}}
\DeclareRobustCommand{\ptsVoxSet}[3]{\ptsSet{#1}{#2}{\vox{}{}{}{}}{#3}}

\newcommand{\cov}[0]{\Lambda}
\DeclareRobustCommand{\pxsCov}[4]{\mat{#1}{#2}{\cov}{#3}{#4}}
\DeclareRobustCommand{\ptsCov}[4]{\mat{#1}{#2}{\dot{\cov}}{#3}{#4}}
\DeclareRobustCommand{\ptsCovSet}[4]{\set{#1}{#2}{\dot{\mathbbold{\cov}}}{#3}{#4}}

\newcommand{\sgm}[0]{\Sigma}
\newcommand{\posesig}[4]{\mat{#1}{#2}{\sgm}{#3}{#4}}

\newcommand{\nbvfunc}[1]{\easyfunc{\eta}{#1}}
\newcommand{\nbsfunc}[1]{\easyfunc{\sigma}{#1}}

\newcommand{\ry}[0]{r}
\DeclareRobustCommand{\ray}[4]{\vtr{#1}{#2}{\ry}{#3}{#4}}
\DeclareRobustCommand{\raySet}[4]{\set{#1}{#2}{\ry}{#3}{#4}}
\DeclareRobustCommand{\rayPos}[3]{\ray{#1}{#2}{\hspace{-0.2em}\pose{}{}}{#3}}
\DeclareRobustCommand{\rayPosSet}[3]{\raySet{#1}{#2}{\hspace{-0.2em}\pose{}{}}{#3}}

\newcommand{\ea}[0]{\lambda}
\DeclareRobustCommand{\eva}[4]{\sca{#1}{#2}{\ea}{#3}{#4}}

\newcommand{\ev}[0]{\nu}
\DeclareRobustCommand{\evt}[4]{\vtr{#1}{#2}{\ev}{#3}{#4}}

\newcommand{\gam}[4]{\sca{#1}{#2}{\gamma}{#3}{#4}}
\newcommand{\alp}[4]{\sca{#1}{#2}{\alpha}{#3}{#4}}
\newcommand{\bng}[4]{\sca{#1}{#2}{\beta}{#3}{#4}}
\newcommand{\prv}[0]{\mysmallmath{PRV}}
\newcommand{\nbv}[0]{\mysmallmath{NBV}}
\newcommand{\nbs}[0]{\mysmallmath{NBS}}
\newcommand{\lft}[0]{V}
\newcommand{\rgt}[0]{S}
\newcommand{\bln}[0]{B}
\newcommand{\ipt}[0]{I}
\newcommand{\ptsint}[0]{\pts{}{}{\mysmallmath{\ipt}}{}}
\newcommand{\gpt}[0]{G}
\newcommand{\voxgeo}[0]{\vox{}{}{\mysmallmath{\gpt}}{}}
\newcommand{\stp}[2]{\sca{}{}{\Delta}{#1}{#2}}

\newcommand{\mylabel}[1]{\ensuremath{#1}}
\newcommand{\occ}[0]{\mylabel{o}}
\newcommand{\emp}[0]{\mylabel{e}}
\newcommand{\uos}[0]{\mylabel{u}}
\newcommand{\wal}[0]{\mylabel{a}}
\newcommand{\dor}[0]{\mylabel{d}}
\newcommand{\wdw}[0]{\mylabel{w}}

\newcommand{\cell}[0]{v}

\newcommand{\mapcell}	[2]	{\mysym{}{}{\cell}{#1}{\mysmallmath{#2}}}
\newcommand{\map}	[2]	{\set{}{}{\cell}{#1}{#2}}

\DeclareRobustCommand{\vox}[4]{\sca{#1}{#2}{\dot{\cell}}{#3}{#4}}
\DeclareRobustCommand{\voxSet}[4]{\set{#1}{#2}{\dot{\cell}}{#3}{#4}}


\newcommand{\fflow}[1]{\easyfunc{flow}{#1}}
\newcommand{\frank}[1]{\easyfunc{rank}{#1}}
\newcommand{\fdiag}[1]{\easyfunc{diag}{#1}}
\newcommand{\funcnn}[1]{\easyfunc{nn}{#1}}

	\newcommand{\dst}[0]{\delta}
	\newcommand{\dist}[2]{\sca{}{}{\dst}{#1}{#2}}
	\newcommand{\distf}[3]{\func{}{}{\dst}{#1}{#2}{#3}}
	\newcommand{\distnn}[1]{\func{}{}{\dst}{nn}{}{#1}}

	\newcommand{\mhl}[0]{h}
	\newcommand{\dmhl}[0]{\sca{}{}{\dst}{\mysmallmath{\mhl}}{}}
	\newcommand{\dmhlf}[1]{\func{}{}{\dst}{\mysmallmath{\mhl}}{}{#1}}
	
\newcommand{\eye}[0]{\mat{}{}{I}{}{}}

\newcommand{\stt}[0]{p}
\newcommand{\state}[4]{\vtr{}{}{\stt}{}{}}
\newcommand{\statespace}[4]{\set{#1}{#2}{\stt}{#3}{#4}}

\newcommand{\cnf}[0]{c}
\newcommand{\conf}[0]{\vtr{}{}{\cnf}{}{}}
\newcommand{\confspace}[4]{\set{#1}{#2}{\cnf}{#3}{#4}}

\newcommand{\rr}[4]{\set{#1}{#2}{R}{#3}{#4}}

\newcommand{\nde}[0]{q}
\newcommand{\node}[4]{\sca{#1}{#2}{\nde}{#3}{#4}}
\newcommand{\nodeset}[4]{\set{#1}{#2}{\nde}{#3}{#4}}
\newcommand{\pth}[0]{\mathscr{T}}
\newcommand{\pathcost}[3]{\func{}{}{\pi}{#1}{#2}{#3}}
\newcommand{\pstps}[0]{Q}
\newcommand{\pathsteps}[0]{\mylabel{\mysmallmath{\pstps}}}

\newcommand{\npart}[0]{N}

\newcommand{\wt}[0]{w}
\newcommand{\wht}	[2]	{\mysym{}{}{\wt}{#1}{#2}}
\newcommand{\whtall}	[2]	{\set{}{}{\wt}{#1}{#2}}
\newcommand{\pa}[0]{s}
\newcommand{\prt}	[2]	{\mysym{}{}{\pa}{#1}{#2}}
\newcommand{\prtall}	[2]	{\set{}{}{\pa}{#1}{#2}}
\newcommand{\os}[0]{z}
\newcommand{\obs}	[2]	{\mysym{}{}{\os}{#1}{#2}}
\newcommand{\obsall}	[2]	{\set{}{}{\os}{#1}{#2}}
\newcommand{\om}[0]{u}
\newcommand{\odom}	[2]	{\mysym{}{}{\om}{#1}{#2}}
\newcommand{\odomCov}	[2]	{\mat{}{}{\Upsilon}{#1}{#2}}
\newcommand{\odomall}	[2]	{\set{}{}{\om}{#1}{#2}}

	\newcommand{\lbl}[2]{\mysym{}{}{\mylabel{\ell}}{#1}{#2}}
	\newcommand{\brn}[2]{\mysym{}{}{\mybearing}{#1}{#2}}
	\newcommand{\rng}[2]{\mysym{}{}{\myrange}{#1}{#2}}
	
	\newcommand{\sig}[2]{\mysym{}{}{\sigma}{#1}{#2}}


\newcommand{\mydepth}[2]{\mysym{}{}{d}{#1}{#2}}
\newcommand{\mybearing}[0]{\ensuremath{\theta}}
\newcommand{\myrange}[0]{\ensuremath{r}}

\newcommand{\setdef}[4]{\mybra{ #1; #2 = #3..#4}}
\newcommand{\setdefsimple}[2]{\setdef{#1}{#2}{1}{\uppercase{#2}}}
\newcommand{\setoftime}[3]{\setdef{#1}{#2}{#3}{#2}}
\newcommand{\setofindex}[4]{\setdef{#1}{#2}{#3}{#4}}
\newcommand{\rayUnitVec}[1]{\ensuremath{\hat{\boldsymbol \theta}_{#1}}}			

\newcommand{\setpreddef}[2]{\mybra{#1 \middle| #2}}

\newcommand{\condprob}[2]{\ensuremath{\myprob{ #1 \middle| #2 }{}}}
\newcommand{\condprobsub}[3]{\ensuremath{\myprob{ #1 \middle| #2 }{#3}}}

\newcommand{\coordframe}[3]{\ensuremath{\prescript{#1}{#2}{#3}}}

\newcommand{\transform}[2]{\tra{\mysmallmath{#2}}{}{\mysmallmath{#1}}{}}
\newcommand{\transformWorld}[1]{\tra{w}{}{\mysmallmath{#1}}{}}
\begin{acronym}

\acro{worldcf}[\coordframe{w}{}{}]{World Coordinate Frame.}

	\acro{rayfunc}[\easyfunc{\rho}{\acs{raypos},\acs{ptsvox}}]{The cost-function for \ac{nbv}.} 
	\acro{voxfunc}[\easyfunc{\upsilon}{\acs{raypos},\acs{ptsvoxset}}]{The cost-function for \ac{nbv}.} 
	\acro{posfunc}[\nbvfunc{\acs{pose}, \acs{ptsall}}]{The cost-function for \ac{nbv}.}
	\acro{nbvfunc}[\nbvfunc{\acs{pose}, \acs{ptsall}}]{The cost-function for \ac{nbv}.} 
	\acro{nbsfunc}[\nbsfunc{\acs{posenbv},\acs{posenbsset},\acs{ptsall}}]{The cost-function for \ac{nbs}.} 

	\acro{bln}[\bln]{}
	\acro{bng}[\bng{}{}{}{}]{}

	\acro{alp}[\alp{}{}{}{}]{}
	\acro{gam}[\gam{}{}{}{}]{}

	\acro{dli}[\dist{\mysmallmath{\lft}\mysmallmath{\ipt}}{}]{}
	\acro{dri}[\dist{\mysmallmath{\rgt}\mysmallmath{\ipt}}{}]{}
	\acro{dba}[\dist{\mysmallmath{\bln}}{}]{}

	\acro{rayl}[\ray{}{}{\mysmallmath{\lft}}{}]{}
	\acro{rayli}[\ray{}{}{\mysmallmath{\lft}\mysmallmath{\ipt}}{}]{}
	\acro{raylg}[\ray{}{}{\mysmallmath{\lft}\mysmallmath{\gpt}}{}]{}
	\acro{rayr}[\ray{}{}{\mysmallmath{\rgt}}{}]{}
	\acro{rayri}[\ray{}{}{\mysmallmath{\rgt}\mysmallmath{\ipt}}{}]{}
	\acro{rayrg}[\ray{}{}{\mysmallmath{\rgt}\mysmallmath{\gpt}}{}]{}

	\acro{poseleft}[\pose{\mysmallmath{\nbv}}{}]{The 2D co-ordinates of a point.}
	\acro{poseright}[\pose{\mysmallmath{\nbs}}{}]{The 2D co-ordinates of a point.}
	
	\acro{posenbv}[\pose{\mysmallmath{NBV}}{}]{The Next-Best View pose.}
	\acro{posenbs}[\pose{\mysmallmath{NBS}}{}]{The Next-Best View pose.}
	\acro{posenbsset}[\poseSet{s}{}]{Set of all possible camera poses.}
	
	\acro{nvswhtbas}[\ensuremath{\wht{\mysmallmath{\bln}}{}}]{}
	\acro{nvswhttri}[\ensuremath{\wht{\mysmallmath{T}}{}}]{}
	\acro{nvswhtrot}[\ensuremath{\wht{\mysmallmath{R}}{}}]{}
	\acro{nvswhtgeo}[\ensuremath{\wht{\mysmallmath{\gpt}}{}}]{}
	\acro{nvscostbas}[\ensuremath{C_{\mysmallmath{\bln}}}]{}
	\acro{nvscosttri}[\ensuremath{C_{\mysmallmath{T}}}]{}
	\acro{nvscostrot}[\ensuremath{C_{\mysmallmath{R}}}]{}
	\acro{nvscostgeo}[\ensuremath{C_{\mysmallmath{\gpt}}}]{}
	
	\acro{gravec}[\vtr{}{}{g}{v}{}]{The gravity vector.}
	\acro{camrotleft}[\mat{}{}{R}{\mysmallmath{\lft}}{}]{The gravity vector.}
	\acro{camrotright}[\mat{}{}{R}{\mysmallmath{\rgt}}{}]{The gravity vector.}
	\acro{pose}[\pose{}{}]{The 3D world co-ordinates of a robot.}
	\acro{posetime}[\pose{\mytime}{}]{The 3D world co-ordinates of a robot at time \mytime.}
	\acro{poseworld}[\poseWorld{}{}]{The 3D world coordinates of a robot in \acs{worldcf}.}

	\acro{poseall}[\poseSet{}{}]{Set of all possible camera poses.}
	\acro{poseallnow}[\poseSet{\mytime}{}]{Set of all camera poses at time \mytime.}

	\acro{pts}[\pts{}{}{}{}]{The 3D co-ordinates of a point.}
	\acro{ptstime}[\pts{}{}{\mytime}{}]{The 3D co-ordinates of a point at time \mytime.}
	\acro{ptsworld}[\pts{}{w}{}{}]{The 3D co-ordinates of a point in \acs{worldcf}.}

	\acro{ptsall}[\ptsSet{}{}{}{}]{Set of all 3D points.}
	\acro{ptsallnow}[\ptsSet{}{}{\mytime}{}]{Set of all camera poses at time \mytime.}

	\acro{ptscov}[\ptsCov{}{}{\pts{}{}{}{}}{}]{Covariance of the point \acs{pts}.}
	\acro{ptscovset}[\ptsCovSet{}{}{}{}]{Set of Covariance matrices.}
	
	\acro{ptsnew}[\pts{}{}{}{\prime}]{The putative match for \acs{pts}.}
	\acro{ptsnewall}[\ptsSet{}{}{}{\prime}]{Set of new 3D points.}
	\acro{ptsnewcov}[\ptsCov{}{}{}{\prime}]{Covariance of the point \acs{pts}.}
	\acro{ptsnewcovset}[\ptsCovSet{}{}{}{\prime}]{Set of Covariance matrices.}
	
	\acro{ptsput}[\pts{}{}{}{\star}]{The putative match for \acs{pts}.}	
	\acro{ptsputcov}[\ptsCov{}{}{}{\star}]{The covariance of the putative match \acs{ptsput}.}
	\acro{ptsputvox}[\pts{}{}{\vox{}{}{}{}}{\star}]{The putative match for \acs{pts} in \vox{}{}{}{}.}	
	\acro{ptsputvoxcov}[\ptsCov{}{}{\vox{}{}{}{}}{\star}]{The covariance of the putative match \acs{ptsputvox}.}
	
	\acro{pxs}[\pxs{}{}{}{}]{The 2D co-ordinates of a point.}
	\acro{pxsprime}[\pxs{}{}{}{\prime}]{The 2D co-ordinates of a point.}
	\acro{pxsleft}[\pxs{}{}{\mysmallmath{\prv}}{}]{The 2D co-ordinates of a point.}
	\acro{pxsright}[\pxs{}{}{\mysmallmath{\nbv}}{}]{The 2D co-ordinates of a point.}
	\acro{pxstime}[\pxs{}{}{}{\mytime}]{The 2D co-ordinates of a point at time \mytime.}
	\acro{pxsidx}[\pxs{}{}{}{\iindex}]{The 2D co-ordinates of a point at index \iindex.}
	\acro{pxsidxprime}[\pxs{}{}{}{\iindex\prime}]{The 2D co-ordinates of a point at index \iindex.}
	
	\acro{pxscov}[\pxsCov{}{}{\pxs{}{}{}{}}{}]{Covariance of the point \acs{pxs}.}
	\acro{pxscovprime}[\pxsCov{}{}{\pxs{}{}{}{\prime}}{}]{Covariance of the point \acs{pxs}.}

	\acro{pxsall}[\pxsSet{}{}{}{}]{Set of all 2D points.}
	\acro{pxsallnow}[\pxsSet{}{}{}{\mytime}]{Set of all 2D points at time \mytime.}

	\acro{ray}[\ray{}{}{}{}]{The 3D co-ordinates of a point.}
	\acro{rayset}[\raySet{}{}{}{}]{Set of new 3D points.}
	
	\acro{raypos}[\ray{}{}{\hspace{-0.2em}\pose{}{}}{}]{The 3D co-ordinates of a point.}
	\acro{rayposset}[\raySet{}{}{\hspace{-0.8mm}\pose{}{}}{}]{Set of new 3D points.}
	
	\acro{rayvox}[\vox{}{}{\ray{}{}{}{}}{}]{An occupied voxel in the octree.}
	
	\acro{evapts}[\eva{}{}{\hspace{-0.5mm}\mysmallmath{\pts{}{}{}{}}}{}]{Eigenvalue of the covariance of point \acs{pts}.}
	\acro{evtpts}[\evt{}{}{\hspace{-0.5mm}\mysmallmath{\pts{}{}{}{}}}{}]{Eigenvector of the covariance of point \acs{pts}.}

	\acro{imgleft}[\img{}{}{\mysmallmath{\prv}}{}]{The left image of a stereo pair.}
	\acro{imgright}[\img{}{}{\mysmallmath{\nbv}}{}]{The right image of a stereo pair.}
  
	\acro{prjleft}[\prj{}{}{\mysmallmath{\prv}}{}]{}
	\acro{prjright}[\prj{}{}{\mysmallmath{\nbv}}{}]{}

	\acro{floleft}[\flo{}{}{\mysmallmath{\prv}}{}]{The flow from left to right image.}
	\acro{floright}[\flo{}{}{\mysmallmath{\nbv}}{}]{The flow from right to left image.}
	
	\acro{amat}[\mat{}{}{A}{}{}]{}
	\acro{amatT}[\mat{}{}{A}{}{\top}]{}
	\acro{bmat}[\vtr{}{}{b}{}{}]{}
	\acro{dmat}[\mat{}{}{B}{}{}]{}
	\acro{dmatT}[\mat{}{}{B}{}{\top}]{}
	\acro{covmat}[\mat{}{}{\bar{\cov}}{}{}]{}
	
	\acro{vox}[\vox{}{}{}{}]{A voxel in the octree.}
	\acro{voxocc}[\vox{}{}{\occ}{}]{An occupied voxel in the octree.}
	\acro{voxemp}[\vox{}{}{\emp}{}]{An empty voxel in the octree.}
	\acro{voxuos}[\vox{}{}{\uos}{}]{An unobserved voxel in the octree.}
	\acro{voxoccset}[\voxSet{}{}{\occ}{}]{An occupied voxel in the octree.}
	\acro{voxempset}[\voxSet{}{}{\emp}{}]{An empty voxel in the octree.}
	\acro{voxuosset}[\voxSet{}{}{\uos}{}]{An unobserved voxel in the octree.}
	\acro{voxset}[\voxSet{}{}{}{}]{The set of voxels that defines the octree.}
	
	\acro{ptsvox}[\ptsVox{}{}{}]{A 3D point in voxel \vox{}{}{}{}.}
	\acro{ptsvoxset}[\ptsVoxSet{}{}{}]{Set of all 3D points in voxel \vox{}{}{}{}.}
	
	\acro{ptsvoxcov}[\ptsCov{}{}{\vox{}{}{}{}}{}]{Set of Covariance matrices.}
	\acro{ptsvoxcovset}[\ptsCovSet{}{}{\vox{}{}{}{}}{}]{Set of Covariance matrices.}
	
	\acro{eye}[\mat{}{}{I}{}{}]{}
	\acro{kgain}[\mat{}{}{K}{g}{}]{Estimated Kalmann Gain for a point update.}
	
	\acro{thrflow}[\thr{f}]{Threshold on the bi-directional optical flow.}
	
	\acro{voxneigh}[\ensuremath{N}]{Number of neighbours in the voxel.}
	
	\acro {funcnn}[\funcnn{\pts{}{}{}{},\ptsSet{}{}{}{}}]{}
	\acro{funcnnerr}[\funcnn{\pts{}{}{}{},\ptsSet{}{}{GT}{}}]{}
	\acro{funcnncov}[\funcnn{\pts{}{}{}{},\ptsSet{}{}{R}{}}]{}
	\acro {distnn}[\distnn{\pts{}{}{}{},\ptsSet{}{}{}{}}]{}
	\acro{distnnerr}[\distnn{\pts{}{}{}{},\ptsSet{}{}{GT}{}}]{}
	\acro{distnncov}[\distnn{\pts{}{}{}{},\ptsSet{}{}{R}{}}]{}

	\acro{posestart}[\pose{start}{}]{The 3D world co-ordinates of a robot.}
	\acro{posegoal}[\pose{goal}{}]{The 3D world co-ordinates of a robot.}
	\acro{poseother}[\pose{o}{}]{The 3D world co-ordinates of a robot.}
	
	\acro{pth}[\ensuremath{\pth}]{}
	\acro{path}[\fnc{}{}{\acs{pth}}{}{}]{}
	\acro{pathprime}[\fnc{}{}{\acs{pth}}{}{\prime}]{}
	\acro{pathshort}[\fnc{}{}{\acs{pth}}{}{\star}]{}
	\acro{coll}[\fnc{}{}{\zeta}{}{}]{}
	\acro{costpar}[\pathcost{}{}{\cdot}]{}
	\acro{costpath}[\pathcost{}{}{\acs{path}}]{}
	\acro{costpathprime}[\pathcost{}{}{\acs{pathprime}}]{}
	\acro{costpathsfm}[\pathcost{sfm}{}{\acs{path}}]{}
	\acro{costpathcol}[\pathcost{col}{}{\acs{path}}]{}
	\acro{pathstart}[\fnc{}{}{\acs{pth}}{}{}\mypar{0}]{}
	\acro{pathgoal}[\fnc{}{}{\acs{pth}}{}{}\mypar{1}]{}
	\acro{statespace}[\statespace{}{}{}{}]{State Space}
	\acro{confspace}[\confspace{}{}{}{}]{Configuration Space}
	\acro{freespace}[\confspace{}{}{free}{}]{Free Configuration Space}
	
	\acro{nbvfuncprt}[\nbvfunc{\prt{\mytime}{\iindex}, \acs{obsnow}}]{The cost-function for \ac{nbv}.} 
	
	\acro{posterior}[\condprob{\prt{\mytime}{}}{\obs{\mytime}{}}]{Likelihood of observation}
	\acro{obsprob}[\condprob{\acs{obsnow}}{\prt{\mytime}{\iindex}}]{Likelihood of observation}
	\acro{nxtprob}[\condprob{\prt{\mytime{+1}}{\iindex}}{\prt{\mytime}{\iindex}}]{Likelihood of observation}
	
	\acro{whtidx}[\wht{}{\iindex}]{Weight of particle \ensuremath{i}.}
	\acro{whtnowidx}[\wht{\mytime}{\iindex}]{Weight of particle \ensuremath{i} at time \mytime.}
	
	\acro{prtnow}[\prt{\mytime}{}]{}
	\acro{prtidx}[\prt{}{\iindex}]{The set of Scene Particles}
	\acro{prtnowidx}[\prt{\mytime}{\iindex}]{The set of Scene Particles}
	\acro{prtprevidx}[\prt{\mytime{-1}}{\iindex}]{The set of Scene Particles}
	\acro{prtall}[\prtall{}{}]{The set of Scene Particles}
	\acro{prtallnow}[\prtall{\mytime}{}]{The set of Scene Particles}
	\acro{prtallnext}[\prtall{\mytime{+1}}{}]{The set of Scene Particles}
	\acro{numpart}[\cst{}{}]{}
	
	\acro{prtres}[\prtall{\mytime}{r}]{}
	\acro{prtuni}[\prtall{\mytime}{u}]{}
	\acro{prtpro}[\prtall{\mytime}{p}]{}
	
	\acro{voxpose}[\vox{}{}{}{\pose{}{}}]{An occupied voxel in the octree.}

	  \acro{rrttree}[\nodeset{}{}{}{}]{}
	  \acro{rrtnn}[\nodeset{}{}{nn}{}]{}
	  \acro{rrtnode}[\node{}{}{}{}]{}
	  \acro{rrtnodestart}[\node{}{}{start}{}]{}
	  \acro{rrtnodegoal}[\node{}{}{goal}{}]{}
	  \acro{rrtnoderand}[\node{}{}{rand}{}]{}
	  \acro{rrtnodenear}[\node{}{}{near}{}]{}
	  \acro{rrtnodenew}[\node{}{}{new}{}]{}
	  \acro{rrtnodenn}[\node{}{}{nn}{}]{}


	\acro{posn}[\posn{}{}]{}
	\acro{posnnow}[\posn{\mytime}{}]{The current pose.}
	\acro{posnprime}[\posn{}{\prime}]{}
	\acro{posnall}[\posnSet{\mytime}{}]{}

	\acro{posngtnow}[\posngt{\mytime}{}]{The current pose.}
	\acro{posngtall}[\posngtSet{\mytime}{}]{}
	
	\acro{posngttf}[\transform{\posnSet{}{}}{\posngtSet{}{}}]{}

\acro{transform}[\transform{}{}]{The 3D world co-ordinates of a robot.}
\acro{transformworld}[\transformWorld{}]{The 3D world coordinates of a robot in \acs{worldcf} at time \mytime}
\acro{transformtime}[\transform{\mytime}{}]{The 3D world co-ordinates of a robot.}
\acro{transformworldprev}[\transformWorld{\mytime{-1}}]{The 3D world co-ordinates of a robot in \acs{worldcf} at time \mytime}
\acro{transformworldnow}[\transformWorld{\mytime}]{The 3D world co-ordinates of a robot in \acs{worldcf} at time \mytime}

	\acro{odomall}[\odomall{\mytime}{}]{\odomall{\mytime}{}}
	\acro{odomnow}[\odom{\mytime}{}]{The Odometry reported by the robot}
	\acro{odomprev}[\odom{\mytime{-1}}{}]{The Odometry reported by the robot}
	\acro{odomnowcov}[\odomCov{\mytime}{}]{}
	
	\acro{obsall}[\obsall{\mytime}{}]{The Observations reported by the sensor.}
	\acro{obsnow}[\obs{\mytime}{}]{The Observations reported by the sensor.}
	\acro{obsprev}[\obs{\mytime{-1}}{}]{The Observations reported by the sensor at time \mytime{-1}.}
	\acro{obsnowidx}[\obs{\mytime}{\kindex}]{}
	\acro{obsnowidxstar}[\obs{\mytime}{\kindex{*}}]{}

	\acro{lblnowidx}[\lbl{\mytime}{\kindex}]{}
	
	\acro{distocc}[\dist{\occ}{}]{}
	\acro{distoccsq}[\dist{\occ}{2}]{}
	\acro{distlbl}[\dist{\mysmallscript{\lbl{}{}}}{}]{}
	\acro{distlblsq}[\dist{\mysmallscript{\lbl{}{}}}{2}]{}
		\acro{distwll}[\dist{\wal}{}]{}
		\acro{distdor}[\dist{\dor}{}]{}
		\acro{distwdw}[\dist{\wdw}{}]{}
	
	\acro{brn}[\brn{}{}]{}
	\acro{brnnow}[\brn{\mytime}{}]{}
	\acro{brnnowidx}[\brn{\mytime}{\kindex}]{}
	
	\acro{rng}[\rng{}{}]{}
	\acro{rngnow}[\rng{\mytime}{}]{}
	\acro{rngnowidx}[\rng{\mytime}{\kindex}]{}
	\acro{rngnowidxstar}[\rng{\mytime}{\kindex{*}}]{}
	
	\acro{lbl}[\lbl{}{}]{}
	\acro{lblnow}[\lbl{\mytime}{}]{}
	\acro{lblnowidx}[\lbl{\mytime}{\kindex}]{}
	
	\acro{dph}[\mydepth{}{}]{}
	\acro{dphnow}[\mydepth{\mytime}{}]{}
	\acro{dphnowidx}[\mydepth{\mytime}{\kindex}]{}
	
	\acro{scantuple}[\mytuple{\acs{brnnowidx}, \acs{rngnowidx}}]{}
	\acro{scantupledef}[\ensuremath{\acs{obsnowidx} = \mytuple{\acs{brnnowidx}, \acs{rngnowidx}}}]{}
	\acro{semtuple}[\mytuple{\acs{brnnowidx}, \acs{rngnowidx}, \acs{lblnowidx}}]{}
	\acro{semtupledef}[\ensuremath{\acs{obsnowidx} = \mytuple{\acs{brnnowidx}, \acs{rngnowidx}, \acs{lblnowidx}}}]{}
	\acro{semtuplenorange}[\mytuple{\acs{brnnowidx}, \acs{lblnowidx}}]{}
	\acro{semtuplenorangedef}[\ensuremath{\acs{obsnowidx} = \mytuple{\acs{brnnowidx}, \acs{lblnowidx}}}]{}
	
	\acro{scanset}[\setdef{\acs{scantuple}}{\kindex}{1}{\Kindex}]{}
	\acro{scanline}[\ensuremath{\acs{obsnow} = \acs{scanset}}]{}
	
	\acro{semset}[\setdef{\acs{semtuple}}{\kindex}{1}{\Kindex}]{}
	\acro{semline}[\ensuremath{\acs{obsnow} = \acs{semset}}]{}

	\acro{map}[\map{}{}]{}
	
	\acro{cell}[\mapcell{\pxs{}{}{}{}}{}]{A cell in the map.}
	\acro{cellat}[\mapcell{\pxs{}{}{}{}}{}]{A cell in the map.}
	\acro{cellatlbl}[\mapcell{\pxs{}{}{}{}}{\lbl{}{}}]{An occupied cell in the map.}
	\acro{cellatocc}[\mapcell{\pxs{}{}{}{}}{\occ}]{An occupied cell in the map.}
	\acro{cellatwdw}[\mapcell{\pxs{}{}{}{}}{\wdw}]{An occupied cell in the map.}
	\acro{cellatdor}[\mapcell{\pxs{}{}{}{}}{\dor}]{An occupied cell in the map.}
	\acro{cellatwal}[\mapcell{\pxs{}{}{}{}}{\wal}]{An occupied cell in the map.}
	
	\acro{cellprt}[\mapcell{\prt{}{}}{}]{A cell in the map.}
	\acro{cellprtlbl}[\mapcell{\prt{}{}}{\lbl{}{}}]{An occupied cell in the map.}
	\acro{cellprtocc}[\mapcell{\prt{}{}}{\occ}]{An occupied cell in the map.}
	\acro{cellprtwdw}[\mapcell{\prt{}{}}{\wdw}]{An occupied cell in the map.}
	\acro{cellprtdor}[\mapcell{\prt{}{}}{\dor}]{An occupied cell in the map.}
	\acro{cellprtwal}[\mapcell{\prt{}{}}{\wal}]{An occupied cell in the map.}
	\acro{cellprtoccnow}[\mapcell{\prt{\mytime}{}}{\occ}]{An occupied cell in the map.}
	\acro{cellprtoccprev}[\mapcell{\prt{\mytime{-1}}{}}{\occ}]{An occupied cell in the map.}
	
	\acro{mapocc}[\map{}{\occ}]{}
	\acro{mapemp}[\map{}{\emp}]{}
	\acro{mapdef}[\ensuremath{\acs{map} = \mypar{\acs{cellocc}, \acs{cellemp}}}]{}
		\acro{cellocc}[\mapcell{}{\occ}]{An occupied cell in the map.}
		\acro{cellemp}[\mapcell{}{\emp}]{An empty cell in the map.}
		\acro{cellatoccnow}[\mapcell{\pxs{}{}{\mytime}{}}{\occ}]{An occupied cell in the map.}
		\acro{cellatoccprime}[\mapcell{\pxs{}{}{}{\prime}}{\occ}]{An occupied cell in the map.}
		\acro{cellatlblprime}[\mapcell{\pxs{}{}{}{\prime}}{\lbl{}{}}]{An occupied cell in the map.}
		\acro{cellatemp}[\mapcell{\pxs{}{}{\mytime}{}}{\emp}]{An empty cell in the map.}
		\acro{cellatuno}[\mapcell{\pxs{}{}{\mytime}{}}{u}]{An unoccupied cell in the map.}
	
	\acro{prtnowprime}[\prt{\mytime}{\iindex\prime}]{}
	\acro{prtallprev}[\prtall{\mytime{-1}}{}]{The set of Scene Particles}
	\acro{prtallnowprime}[\prtall{\mytime}{\prime}]{The set of Scene Particles}
	\acro{prtallprevprime}[\prtall{\mytime{-1}}{\prime}]{The set of Scene Particles}
 	\acro{prtprev}[\prt{\mytime{-1}}{\iindex}]{Current Particle}
 	\acro{prtprevprime}[\prt{\mytime{-1}}{\iindex{\prime}}]{Current Particle}
	
	\acro{whtnow}[\wht{\mytime}{}]{Weights at time \mytime.}
	\acro{whtnowprime}[\wht{\mytime}{\iindex{\prime}}]{Weight of particle \ensuremath{i} at time \mytime.}
	\acro{whtprev}[\wht{\mytime{-1}}{\iindex}]{Weight of particle \ensuremath{i} at time \mytime{-1}.}
	
	\acro{sighit}[\sig{\occ}{}]{}
	\acro{sighitsq}[\sig{\occ}{2}]{}
	
	\acro{siglbl}[\sig{\mysmallscript{\lbl{}{}}}{}]{}
	\acro{siglblsq}[\sig{\mysmallscript{\lbl{}{}}}{2}]{}
	
	\acro{zhit}[\myexp{\slfrac{{-\mypar{\rng{\mytime}{\kindex}-\rng{\mytime}{\kindex{*}}}^2}}{{2\sig{\occ}{2}}}}]{}
	\acro{zhit_field}[\myexp{\slfrac{{-\dist{\mysmallscript{\occ}}    {2}}}{{2\sig{\mysmallscript{\occ}    }{2}}}}]{}
	\acro{lhit_field}[\myexp{\slfrac{{-\dist{\mysmallscript{\lbl{}{}}}{2}}}{{2\sig{\mysmallscript{\lbl{}{}}}{2}}}}]{}
	
	\acro{lasergaus}[\mygaussian{\acs{rngnowidx};\acs{rngnowidxstar}}{\acs{sighitsq}}]{}
	\acro{lasergausdef}[\mygaussian{\acs{rngnowidx};\acs{rngnowidxstar}}{\acs{sighitsq}} = \acs{zhit}]{}



	\acro{probmclall}[\condprob{\acs{prtnowidx}}{\acs{obsall},\acs{odomall}}]{}
	\acro{probmcl}[\condprob{\acs{prtnowidx}}{\acs{obsnow},\acs{odomnow}}]{}
	\acro{prior}[\condprob{\acs{prtprev}}{\acs{obsprev},\acs{odomprev}}]{}
	\acro{probmotionprior}[\condprob{\acs{prtprevidx}}{\acs{map}}]{}
	\acro{probmotion}[\condprob{\acs{prtnowprime}}{\acs{odomnow}, \acs{prtprev}}]{}
	\acro{probmotionmap}[\condprob{\acs{prtnowprime}}{\acs{odomnow}, \acs{prtprev}, \acs{map}}]{}
	\acro{probsensor}[\condprob{\acs{obsnow}}{\acs{prtnowprime}, \acs{map}}]{}

	\acro{probray}[\condprob{\acs{obsnowidx}}{\acs{prtnowprime}, \acs{map}}]{}
	
	\acro{priorlbl}[\myprob{\lbl{}{}}{}]{}
	
		\acro{ghtfac}[\ensuremath{\epsilon_{\mysmallscript{g}}}]{}
		\acro{lblfac}[\ensuremath{\epsilon_{\lbl{}{}}}]{}
		
			\acro{whtrng}[\ensuremath{\epsilon_{\mysmallmath{\occ}}}]{}
			\acro{whtlbl}[\ensuremath{\epsilon_{\mysmallmath{\lbl{}{}}}}]{}
		
			\acro{whthit}[\ensuremath{\epsilon_{\mysmallscript{hit}}}]{}
			\acro{whtsrt}[\ensuremath{\epsilon_{\mysmallscript{short}}}]{}
			\acro{whtmax}[\ensuremath{\epsilon_{\mysmallscript{max}}}]{}
			\acro{whtrnd}[\ensuremath{\epsilon_{\mysmallscript{rand}}}]{}
	
	\acro{probrng}[\condprobsub{\acs{obsnowidx}}{\acs{prtnowprime}, \acs{map}}{\mysmallscript{rng}}]{}
	\acro{problbl}[\condprobsub{\acs{obsnowidx}}{\acs{prtnowprime}, \acs{map}}{\mysmallscript{lbl}}]{}
	
	\acro{probhit}[\condprobsub{\acs{obsnowidx}}{\acs{prtnowprime}, \acs{map}}{\mysmallscript{hit}}]{}
	\acro{probshort}[\condprobsub{\acs{obsnowidx}}{\acs{prtnowprime}, \acs{map}}{\mysmallscript{short}}]{}
	\acro{probmax}[\condprobsub{\acs{obsnowidx}}{\acs{prtnowprime}, \acs{map}}{\mysmallscript{max}}]{}
	\acro{probrand}[\condprobsub{\acs{obsnowidx}}{\acs{prtnowprime}, \acs{map}}{\mysmallscript{rand}}]{}



\end{acronym}

\maketitle
\thispagestyle{empty}
\pagestyle{empty}
\begin{abstract}

How does a person work out their location using a floorplan?
It is probably safe to say that we do not explicitly measure depths to every visible surface and try to match them against different pose estimates in the floorplan. 
And yet, this is exactly how most robotic scan-matching algorithms operate. 
Similarly, we do not extrude the 2D geometry present in the floorplan into 3D and try to align it to the real-world. 
And yet, this is how most vision-based approaches localise.

Humans do the exact opposite.
Instead of depth, we use high level semantic cues.
Instead of extruding the floorplan up into the third dimension, we collapse the 3D world into a 2D representation.
Evidence of this is that many of the floorplans we use in everyday life are not accurate, opting instead for high levels of discriminative landmarks.

In this work, we use this insight to present a global localisation approach that relies solely on the semantic labels present in the floorplan and extracted from RGB images.
While our approach is able to use range measurements if available, we demonstrate that they are unnecessary as we can achieve results comparable to state-of-the-art without them.

\end{abstract}

\vspace{-0.1cm}
\section{Introduction}
\vspace{-0.2cm}
Indoor localisation is perhaps one of the most crucial aspects for any robotic system. %
It allows robots to interact with the world and provides a representation and understanding that can be shared with humans and other agents.
Traditional \ac{vslam} systems can provide localisation within a map that is built on-the-fly.
However, \ac{vslam} systems are liable to drift in terms of both pose and scale. 
They can also become globally inconsistent in the case of failed loop closures.
Finally, even in the case of no scale drift and correct loop closures, a \ac{vslam} system can only ever guarantee global consistency \textit{internally}.
This means that while pose estimates are globally consistent, they are only valid within the context of the \ac{vslam} system.
There are no guarantees, at least in vision-only systems, that we can directly map the reconstruction to the real world (or between agents).

\begin{figure}
\vspace{-0.5cm}
  \includegraphics[width=1.0\linewidth]{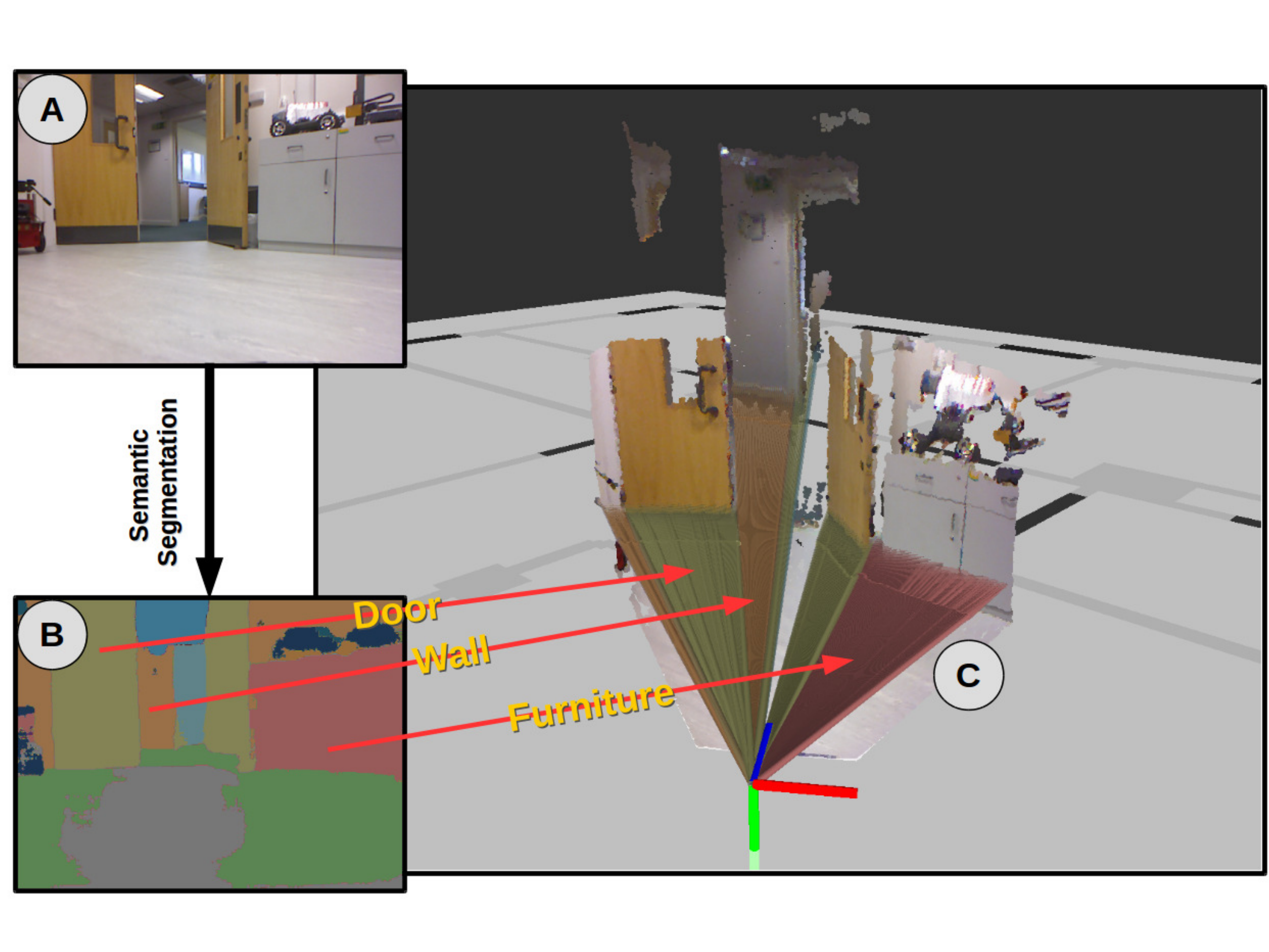}
\vspace{-0.8cm}
  \caption{A) RGB Image, B) CNN-Based Semantic Labelling and C) Sample \acs{sedar} Scan within floorplan.}  
\vspace{-0.8cm}
  \label{sec:intro:fig:sedar}
\end{figure}

This problem is normally addressed by having a localisation system that can relate the pose of the robot to a pre-existing map.
Examples of global localisation frameworks include the \ac{gps} and traditional \ac{mcl}.
\ac{mcl} has the ability to localise within an existing floorplan (which can be safely assumed to be available for most indoor scenarios).
This is a highly desirable trait, as it implicitly eliminates drift, is globally consistent and provides a way for the created 3D reconstructions to be related to the real world without having to perform expensive post-hoc optimizations.
Traditionally, the range-based scans required by \ac{mcl} have been produced by expensive sensors such as \ac{lidar}.
More recently, modern robotic platforms have used RGB-D cameras as a cheap and low-footprint alternative.
This has made vision-based floorplan localisation an active topic in the literature.

However, while several vision-based approaches have been proposed, they normally use heuristics to lift the 2D plan into the 3D coordinate system of \ac{vslam}. 
Examples include Liu \etal \cite{Liu2015a}, who use visual cues such as \acp{vp} or Chu \etal \cite{Chu2016} who perform piecemeal 3D reconstructions that can then be fitted back to an extruded floorplan.
A common problem with these approaches is that the 3D data extracted from the image is normally orthogonal to the floorplan that it is meant to localise in.
This means that assumptions must be made about dimensions not present in the floorplan.
These approaches also do not fully exploit the floorplan, ignoring the semantic information.

We propose a fundamentally different approach that is inspired by how humans perform the task.
Instead of discarding valuable semantic information, we use a \ac{cnn}-based encoder-decoder to extract high-level semantic information. 
We then collapse all semantic information into 2D in order to reduce the assumptions about the environment.
We then use these labels, image geometry and (optionally) depth along with a semantically labelled floorplan to create a state-of-the-art sensing and localisation framework.\looseness=-1

\acf{sedar} is an innovative human-inspired framework that combines new semantic sensing capabilities with a novel semantic \acf{mcl} approach.
As an example, figure \ref{sec:intro:fig:sedar} shows a sample \ac{sedar} scan localised in the floorplan.
We show that \ac{sedar} has the ability to surpass \ac{lidar}-based \ac{mcl} approaches.
\ac{sedar} also has the ability to perform drift-free local, as well as global, localisation.
Furthermore, experimental results show that the semantic labels are sufficiently strong visual cues such that depth estimates are no longer needed.
Not only does this vision-only approach perform comparably to depth-based methods, it is also capable of coping with map inaccuracies more gracefully.

This paper describes the process by which \ac{sedar} is used as a novel human-inspired sensing and localisation framework.
In section \ref{sec:meth:floorplan}, semantically salient elements are extracted from a floorplan.
Section \ref{sec:meth:semseg} describes how these semantic elements are identified in the robot's camera by using a state-of-the-art \ac{cnn}-based semantic segmentation algorithm and presented as a novel sensing modality.
We then present the three main contributions of this paper.
First, section \ref{sec:meth:mcl:mm} introduces a novel motion model that includes a ``ghost factor'' that uses semantic information to influence how particles move through occupied space.
Second, section \ref{sec:meth:mcl:sm} introduces a novel sensor model that estimates observation likelihoods using semantic information, range and bearing information.
Third, section \ref{sec:meth:mcl:sm:lbl} introduces a second novel motion model that uses semantic and bearing information to allow observation likelihoods to be estimated from an RGB image only.
Finally, in section \ref{sec:res} we present the results obtained by using our approach in multiple sensing modalities.

\vspace{-0.1cm}
\section{Literature Review}\label{sec:lit}
\vspace{-0.1cm}
\acf{mcl} was made possible by the arrival of accurate range-based sensors such as \ac{sonar} and \acf{lidar}.
These approaches, which we call \acf{rmcl}, are robust and reliable and still considered state-of-the-art in many robotic applications.
Recent advances in computer vision have made it possible for us to imagine new types of perceptual sensors which are capable of semantic understanding of a scene.
Semantic sensing modalities, such as \ac{sedar}, have the ability to revolutionize \ac{mcl}. %

\ac{rmcl} was first introduced by Fox \etal \cite{Fox1999a} and Dellaert \etal \cite{Dellaert1999}.
\ac{rmcl} improved the Kalman Filter based state-of-the-art by allowing multi-modal distributions to be represented.
It also solved the computational complexity of grid-based Markov approaches.
However, these approaches require expensive \ac{lidar} and/or \ac{sonar} sensors to operate reliably.
Instead, Dellaert \etal \cite{Dellaert1999a} extended their approach to use vision-based sensor models.
Vision-based \ac{mcl} allowed the use of rich visual features and cheap sensors, but had limited performance compared to the more robust \ac{lidar}-based systems.\looseness=-1

With the rising popularity of RGB-D sensors, more robust vision-based \ac{mcl} approaches became possible.
Paton and Kosecka \cite{Paton2012} use a combination of feature matching and \ac{icp} to perform pose estimation and localisation.
Brubaker \etal \cite{Brubaker2013} used visual odometry and pre-existing roadmaps in a joint \ac{mcl}/closed-form approach in order to localise a moving car.
Fallon \etal \cite{Fallon2012a} presented a robust \ac{mcl} approach that used a low fidelity \textit{a priori} map to localise in, but required the space to be traversed by a depth sensor beforehand.
Winterhalter \etal \cite{Winterhalter2015} performed \ac{mcl}, but based the likelihood of the sensor model on the normals of an extruded floorplan.
Chu \etal \cite{Chu2016} is the closest to us, they attempted to mimic the human thinking process by creating piecemeal reconstructions of an extruded floorplan, the \ac{mcl} sensor model was then based on matches against these reconstructions.
These \ac{mcl}-based approaches tend to be robust, but they operate entirely on the \textit{geometric} information present in the floorplan and therefore require depth images either from sensors and/or reconstructions. 
By contrast our approach aims to use \textit{non-geometric} semantic information present in the floorplan in order to perform the localisation.\looseness=-1

Our approach is most similar to bearing-only \cite{Briechle2004,Thrun2002} approaches, where the angular distrubtion of known landmarks can be used to deduce the location of a robotic agent.
However, our approach is fundamentally different from these methods, as it does not require active landmarks with known positions.
Instead, we rely on the semantic information already present in the world: we use the angular distribution of detected semantic labels to localise a robot.

While the field of \ac{mcl} evolved in the robotics community, in vision, the non-\acs{mcl}-based field of floorplan localisation became more popular.
Melbouci \etal \cite{Melbouci2016} used extruded floorplans, but performed local bundle adjustments instead of \ac{mcl}. 
Shotton \etal \cite{Shotton2013} used regression forests to predict the correspondences of every pixel in the image to a known 3D scene, they then combined this in a \ac{ransac} approach in order to solve the camera pose.
Chu \etal \cite{Chu2016a} use information from the floorplans and Google StreetView in order to reason about the geometry of the building and perform a robust reconstruction.
The most similar work to our approach is Wang \etal \cite{Wang2015} who use text detection from shop fronts as semantic cues to localise in the floorplan of a shopping centre and Liu \etal \cite{Liu2015a} who use floorplans as a source of geometric and semantic information, combined with vanishing points, to localise monocular cameras.
These vision-based approaches tend to use more of the non-geometric information present in the floorplan.
However, a common trend is that assumptions must be made about geometry not present in the floorplan (\eg ceiling height).
The floorplan is then extruded out into the \nth{3} dimension to allow approaches to use the information present in the image.
By contrast, our approach aims to extract the information from the image and collapse the 3D world down into the 2D floorplan where localisation can be performed.
This provides a 3-\ac{dof} localisation requiring less assumptions about the environment.

Recently, advances in Deep Learning have made robust semantic segmentation models widely available.
Approaches like that of Badrinarayanan \etal \cite{Badrinarayanan2015}, Kendal \etal \cite{Kendall2015} and Long \etal \cite{Shelhamer2017} have made semantically informed approaches possible.
One such approach is Tateno \etal \cite{Tateno2017} who use the \ac{cnn}-based depth and semantic label predictions of Laina \etal \cite{Laina2016} to aid in their \ac{slam} pipeline.
Lee \etal \cite{Lee2017} extend the approach of Badrinarayanan \etal \cite{Badrinarayanan2015} to directly estimate room layout keypoints.
While many such approaches exist, they mainly focus on extracting the room layout based on Manhattan world assumptions. 
Instead, this work proposes to use \ac{cnn}-based semantic segmentation (that is understandable to humans) in order to extract labels that are inherently present in human-readable floorplans.
This allows us to take all that information and collapse it into a 3-\ac{dof} problem, making our approach more tractable than competing 6-\ac{dof} approaches while avoiding additional assumptions.
\begin{figure*}
  \begin{center}
    \subfloat[Original floorplan]{\includegraphics[width=0.155\linewidth]{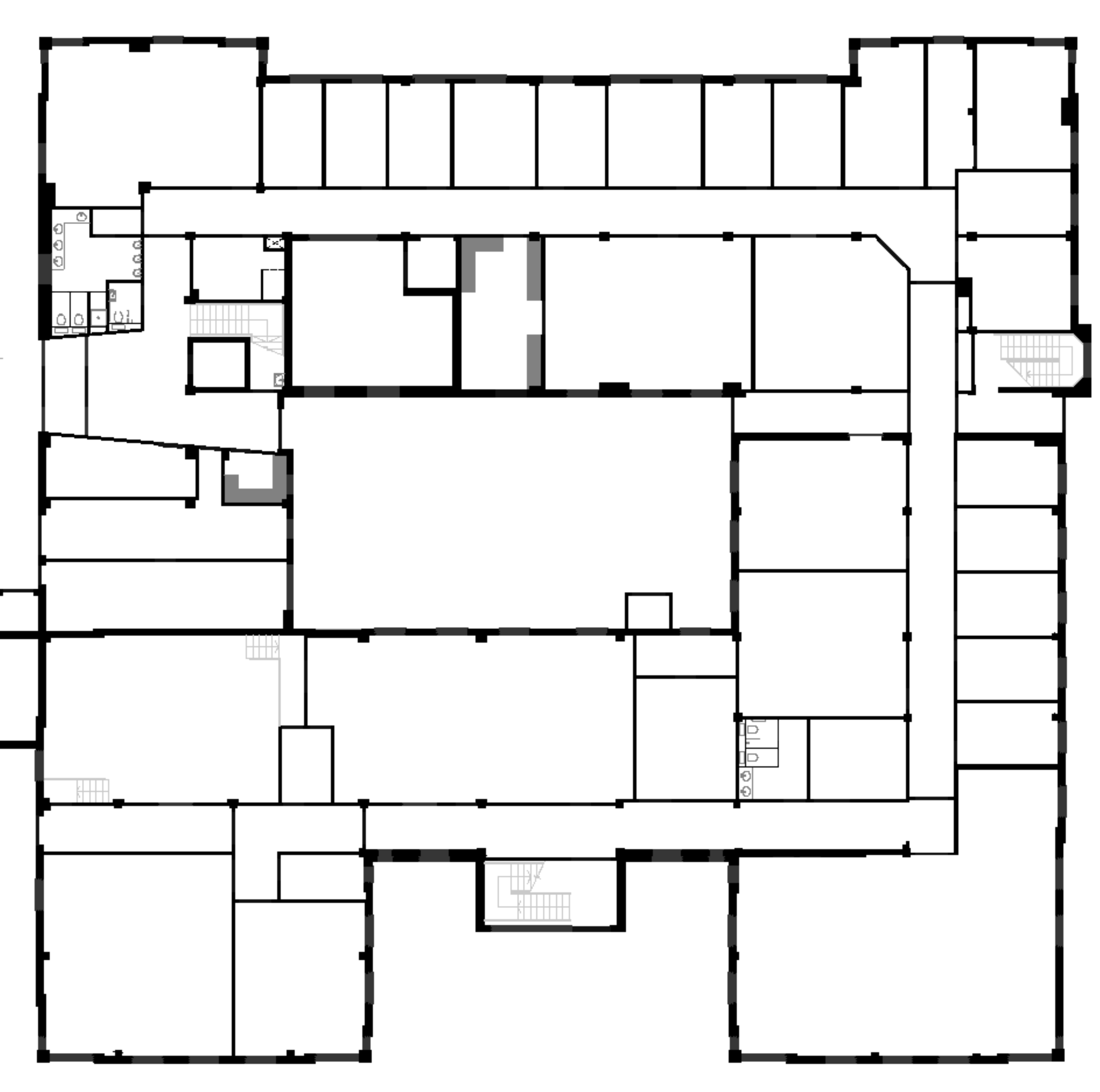}\label{sec:meth:fig:floorplan}}\hspace{0.5mm}
    \subfloat[Occupancy]{\includegraphics[width=0.155\linewidth]{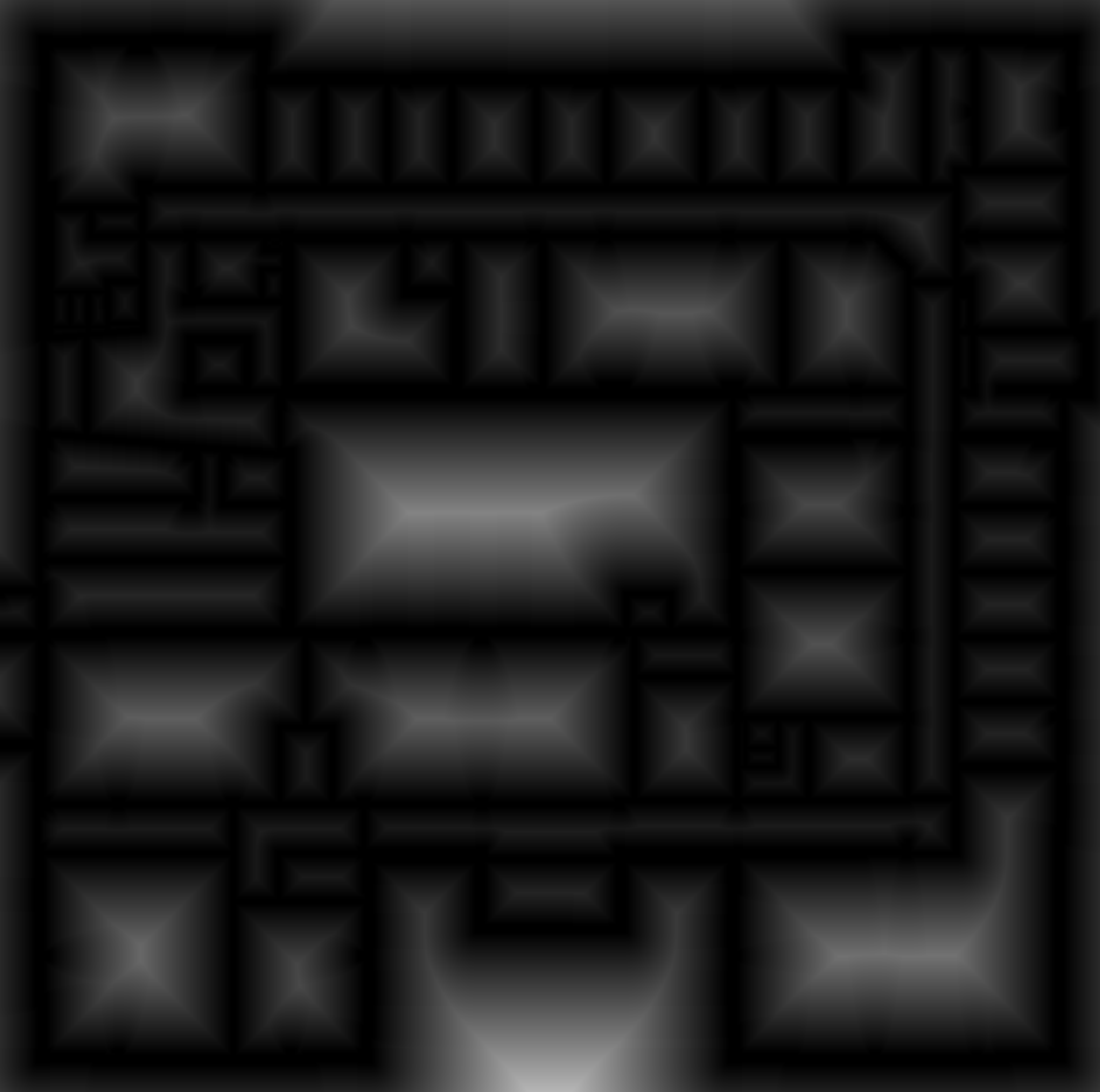}\label{sec:meth:fig:field}}\hspace{0.5mm}
    \subfloat[Semantic Floorplan]{\includegraphics[width=0.155\linewidth]{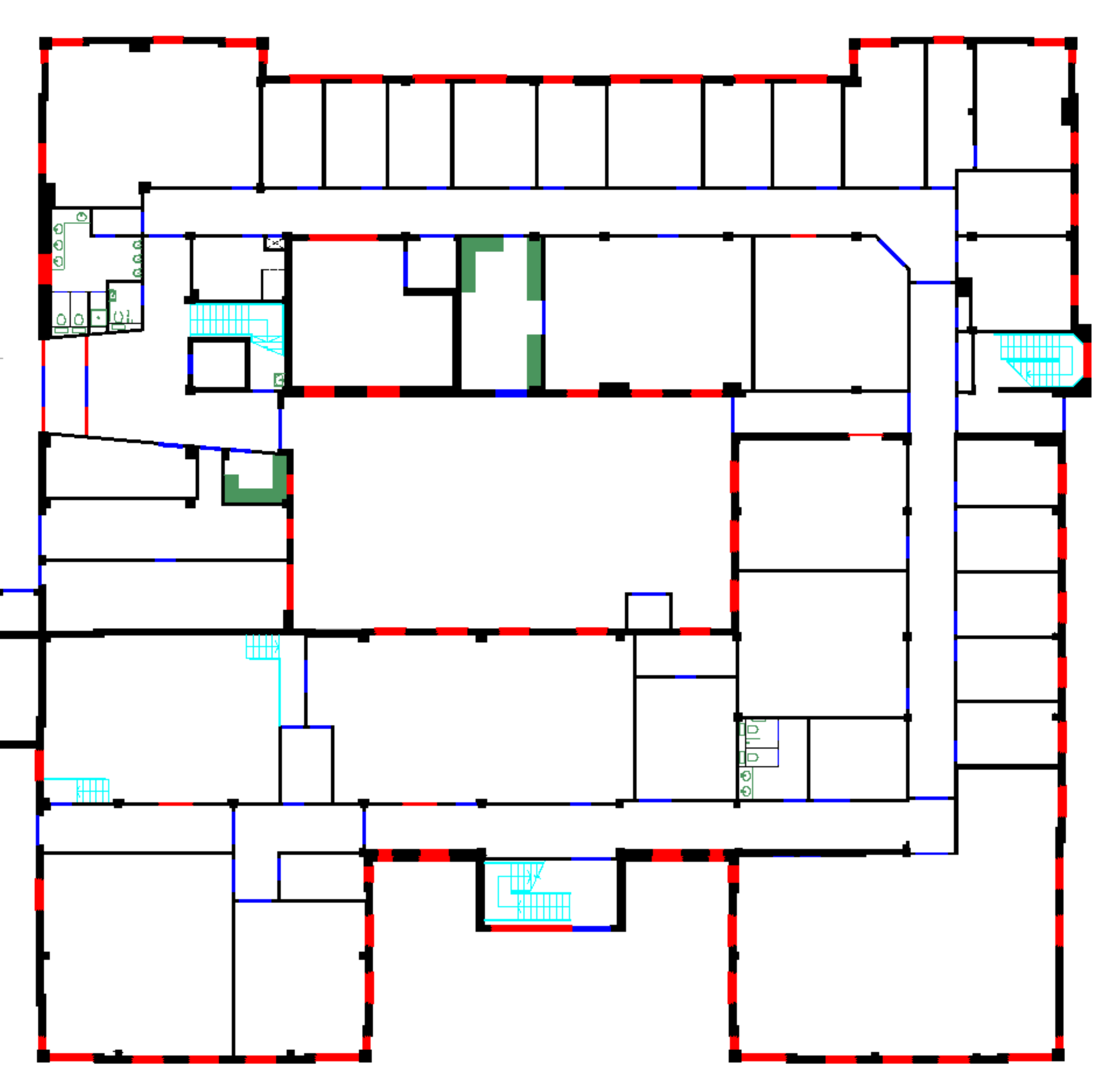}}\hspace{0.5mm}
    \subfloat[Wall Likelihood]{\includegraphics[width=0.155\linewidth]{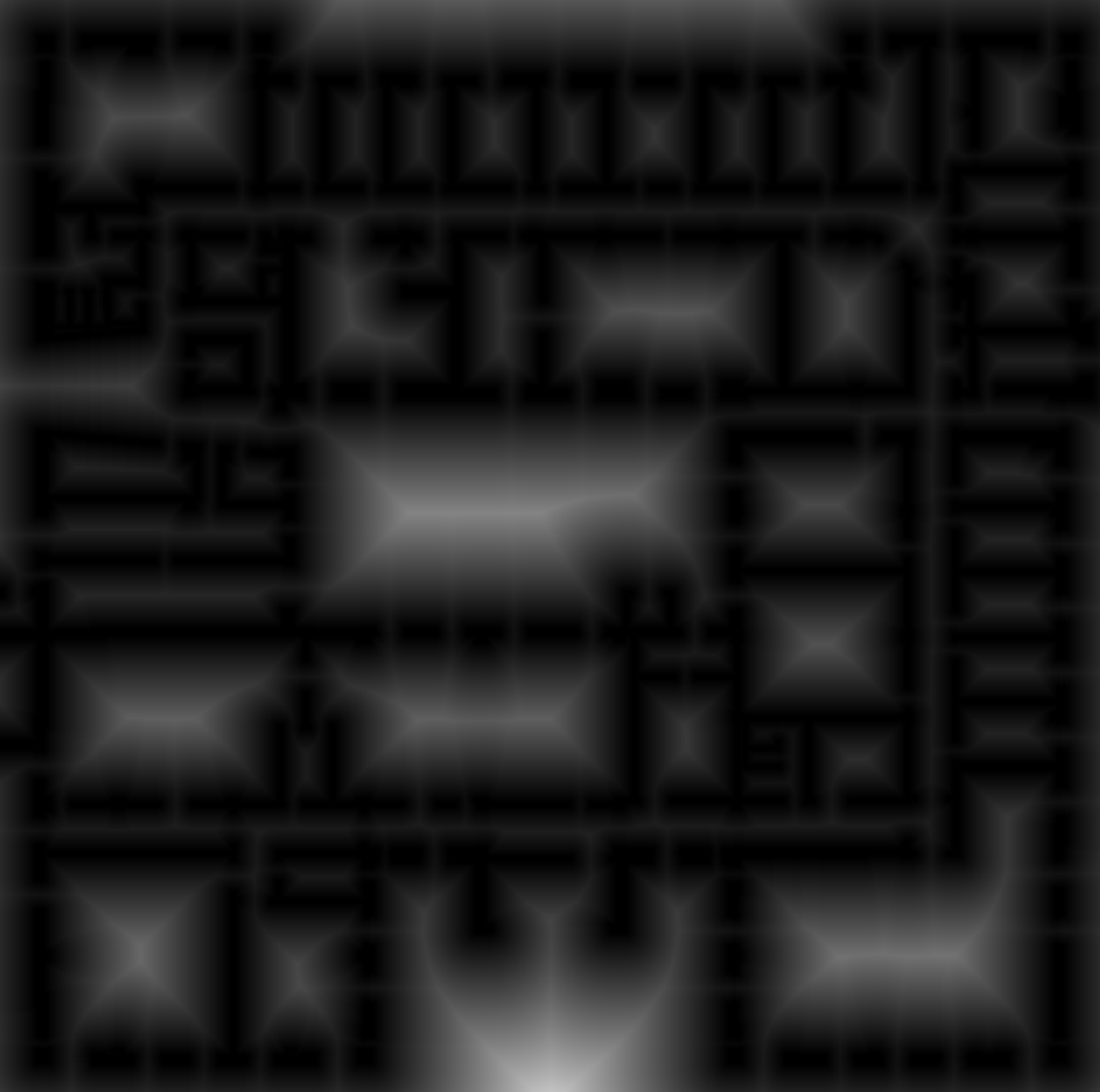}}\hspace{0.5mm}
    \subfloat[Door Likelihood]{\includegraphics[width=0.155\linewidth]{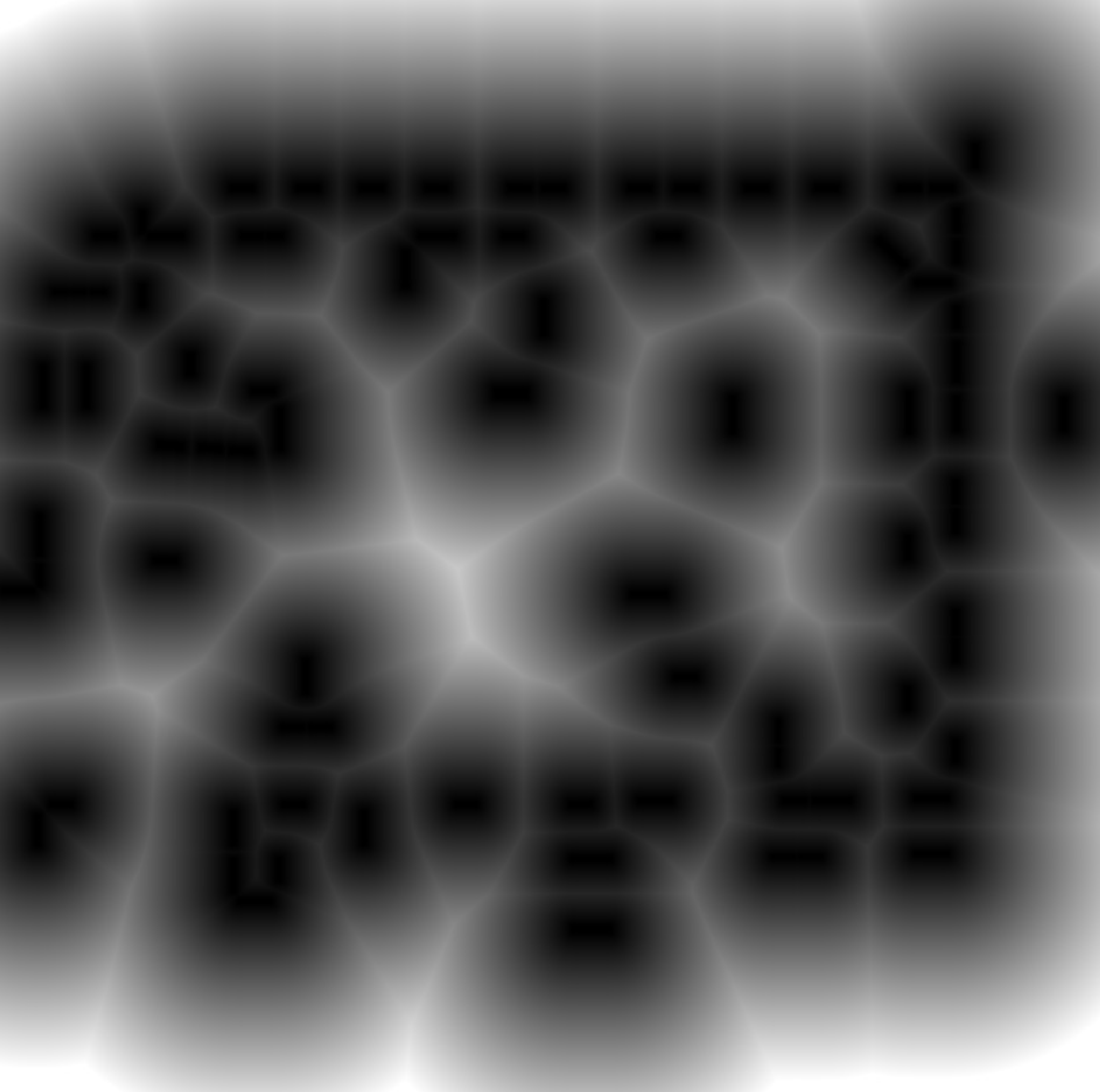}}\hspace{0.5mm}
    \subfloat[Window Likelihood]{\includegraphics[width=0.155\linewidth]{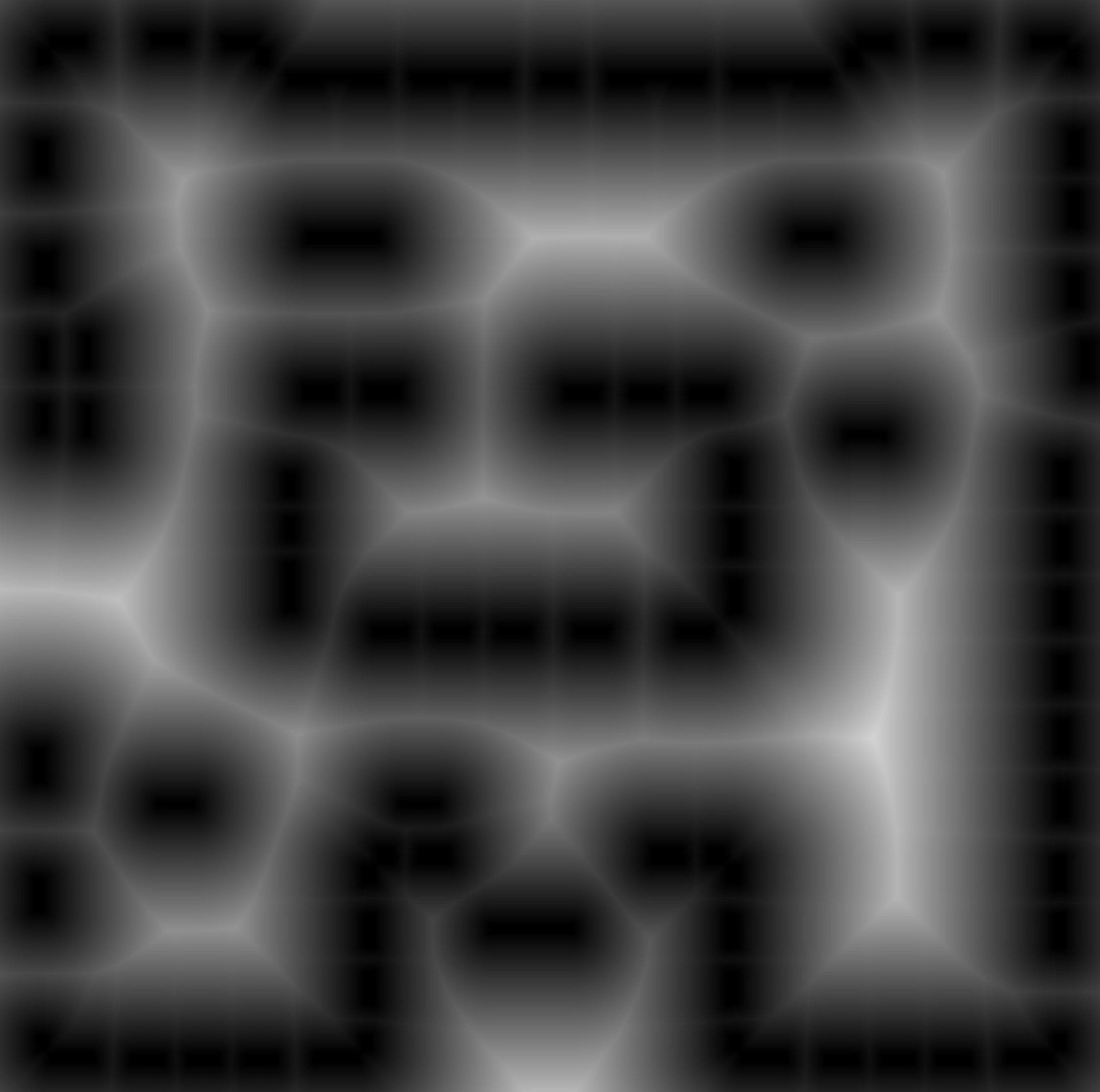}}
  \end{center}  
	\vspace{-0.2cm}
  \caption{Left: Original floorplan and occupancy likelihood field. Right: semantic floorplan and label likelihood fields.}  
\label{fig:meth:floorplans}
\vspace{-0.6cm}
\end{figure*}
\vspace{-0.2cm}
\section{Methodology}\label{sec:meth}
\vspace{-0.2cm}
The problem with state-of-the-art approaches is that they 
are limited to range information. 
Instead, we present a novel semantic sensing and localisation framework called \ac{sedar} that leverages semantic and, optionally, range information.
We will show that we can use our novel \ac{sedar} sensing and localisation framework to outperform traditional \ac{rmcl}.
\vspace{-0.2cm}
\subsection{Semantic Floorplans}\label{sec:meth:floorplan}
\vspace{-0.1cm}
\ac{rmcl} requires a floorplan and/or previously created range-scan map that is accurate in scale and globally consistent.
Use of human-readable floorplans makes a system much more broadly applicable than relying on prior exploration and mapping.
However, differences between the floorplan and the robot's observations (\eg inaccuracies, scale variation and furniture) can reduce the reliability.

To overcome this, we augment the localisation with semantic labels extracted from an existing floorplan.
In our experiments we limit these labels to walls, doors and windows (see figure \ref{fig:meth:floorplans}), which are easy to automatically extract from a floorplan, and are also salient for human localisation.

In order to make a labelled floorplan readable by the robot, it must first be converted into an occupancy grid.
An occupancy grid is a 2D representation of the world, in which each cell in the grid has an occupancy probability, determined by it's normalized greyscale value.

If \ensuremath{\pxsSet{}{}{}{}} is a set of 2D positions, the map can then be defined as 
$\acs{map} = \mybra{\acs{cellat} ; \pxs{}{}{}{} \in \pxsSet{}{}{}{} \subset \mathbb{Z}^{+2}}$.
Then, assuming \ensuremath{\mathcal{L} = \mybra{\wal, \dor, \wdw}} is the set of possible cell labels (wall, door, window), each cell is defined as
$
  \acs{cellat} = \mytuple{\acs{cellatocc}, \acs{cellatwdw}, \acs{cellatdor}, \acs{cellatwal}}
$
where \ensuremath{\acs{cellatocc}} is the occupancy likelihood and \ensuremath{\lbl{}{} \in \mathcal{L}} denotes the label likelihood.
\vspace{-0.2cm}
\subsection{\ac{sedar} Sensor}\label{sec:meth:semseg}
\vspace{-0.1cm}
Modern low-cost robotics systems turn the RGB-D image received at time \ensuremath{\mytime} into a set of range (\ensuremath{\acs{rngnowidx}}) and bearing (\ensuremath{\acs{brnnowidx}}) tuples. %
\ac{sedar} adds a semantic label (\ensuremath{\acs{lblnowidx}}) to this tuple.
Instead of using the whole image simultaneously (which would be intractible), tuples are arranged along horizontal scanlines (\ensuremath{\acs{semline}}), where \ensuremath{\kindex} is the horizontal pixel location.
In this work, the centre scanline is assumed to be parallel with the ground plane and is therefore used to collapse the 3D information of the RGB-D image into the 2D floorplan. 

While range and bearing values can be extracted using simple geometry, their corresponding labels must be estimated using a state-of-the-art semantic segmentation algorithm.
Any semantic segmentation approach can be used, however, Deep Learning based approaches currently dominate the benchmarks \cite{cordts2016} in this field.
Therefore, a \ac{cnn}-based encoder-decoder network \cite{Kendall2015} is used.
This is trained on the SUN3D \cite{Xiao2013} dataset, and can reliably detect doors, walls, floors, ceilings, furniture and windows.
This state-of-the-art semantic segmentation runs in real-time, which allows images to be parsed into a \ac{sedar}-scan with negligible latency.
The label \ensuremath{\acs{lblnowidx}} is then simply the label at pixel \ensuremath{\kindex} along the horizontal scanline. \looseness=-1

It is important to note that we extract the labels from the RGB image only.
This is by design, as it allows the use of cameras that cannot sense depth.
In the following sections we will use this novel sensing modality in a novel \ac{mcl} formulation with and without the range-based measurements.
\vspace{-0.6cm}
\subsection{Motion Model}\label{sec:meth:mcl:mm}
\vspace{-0.1cm}
\ac{mcl} motion models are normally represented by the distribution \ensuremath{\acs{probmotion}}, where the previous set of particles \ensuremath{\acs{prtprevidx}} is propagated using the odometry measurements \ensuremath{\acs{odomnow}} into the current set of particles \ensuremath{\acs{prtnowidx}}.
However, it is well understood in the literature that the actual distribution being approximated is \ensuremath{\ensuremath{\acs{probmotionmap}}}.
This encodes the idea that certain motions are more or less likely depending on the map (\eg through walls).
Under the assumption that the motion of the robot is small, it can be shown that 
\vspace{-0.1cm}
\begin{equation}
\!\! \acs{probmotionmap} \!=\! \kappa \acs{probmotion} \acs{probmotionprior}
\vspace{-0.1cm}
\end{equation}
(see \eg \cite{Thrun2002}) where \ensuremath{\kappa} is a normalising factor and \ensuremath{\acs{map}} is the set containing every cell in the map.
This allows the two likelihoods to be treated independently.
The motion \ensuremath{\acs{probmotion}} is defined as in RCML \cite{Thrun2002}. 
The prior is the occupancy likelihood of the cell that contains \ensuremath{\acs{prtnowidx}}, that is \ensuremath{\acs{probmotionprior} = 1 - \myprob{\acs{cellprtoccprev}}{}}

However, this prior estimation approach becomes problematic when using human-made floorplans, as these typically have image artefacts introduced during the scanning process.
Therefore, most approaches threshold the occupancy
\vspace{-0.1cm}
\begin{equation}
 \myprob{\acs{cellprtoccprev}}{} =
\begin{cases}
    1 &  \mbox{if } \acs{cellprtoccprev} \geq \thr{o}\\
    0 &  \mbox{otherwise}
\end{cases}
\label{eq:sec:meth:occprob}
\vspace{-0.1cm}
\end{equation}
where $\tau_o$ is a user defined threshold.
This exacerbates problems with floorplan accuracy and occlusions.
For instance, most humans would not even notice if a door is a few centimetres away from where it should be.
However, this presents real problems when particles propagate though doors, as many valid particles will be discarded upon contact with the expected edge of the door frame.
Instead, we propose to augment this with a \textit{ghost factor} (\ensuremath{\acs{ghtfac}}) that allows particles more leeway in these scenarios.
Therefore the proposed prior is 
\vspace{-0.1cm}
\begin{equation}
  \acs{probmotionprior} = \mypar{1 - \myprob{\acs{cellprtoccprev}}{}} e^{\minus \acs{ghtfac} \acs{distwll}}
	\vspace{-0.1cm}
\end{equation}
where \ensuremath{\acs{distwll}} is the distance to the nearest door.
While other labels such as windows can be used, in the case of a ground-based robot doors are sufficient.
The distance, \ensuremath{\acs{distwll}}, can be efficiently estimated using a lookup table as defined in section \ref{sec:meth:mcl:sm}.\looseness=-1

More importantly, \ensuremath{\acs{ghtfac}} is a user defined factor that determines how harshly this penalty is applied.
Setting \ensuremath{\acs{ghtfac} = 0} allows particles to navigate through walls with no penalty, while very high values approximate equation \ref{eq:sec:meth:occprob}.
We will explore the effects of \ensuremath{\acs{ghtfac}} in section \ref{sec:res:ght}.
This motion model is more probabilistically accurate than the occupancy model used in most \ac{rmcl} approaches, and has the added advantage of leveraging the high-level semantic information present in the map.\looseness=-1

\vspace{-0.2cm}
\subsection{Sensor Model}\label{sec:meth:mcl:sm}
\vspace{-0.1cm}
The na\"ive way of incorporating semantic measurements into the sensor model would be to use the beam model. 
In this modality, the raycasting operation would provide not only the distance travelled by the ray, but also the label of the cell the ray hit. 
If the label of the cell and the observation match, the likelihood of that particle being correct is increased. 
However, this approach suffers from the same limitations as the traditional beam model: it has a distinct lack of smoothness. 
On the other hand, the likelihood field model is significantly smoother, as it provides a gradient between each of the cells. 
By contrast, the approach presented here uses a joint method that can use likelihood fields to incorporate semantic information in the presence of semantic labels.
More importantly, it can also use raycasting within a likelihood field in order to operate without range measurements.

The likelihood field model calculates a distance map.
For each cell \ensuremath{\acs{cellat}}, the distance to the nearest occupied cell
\vspace{-0.2cm}
\begin{equation}
	\acs{distocc}\mypar{\acs{pxs}} = \min_{\pxs{}{}{}{\prime}}\mymag{\acs{pxs}-\acs{pxsprime}}, \quad \acs{cellatoccprime} > \thr{o}
	\label{eq:meth:distocc}
	\vspace{-0.2cm}
\end{equation}
is calculated and stored. 
For clarity, we omit the parameter \ensuremath{\acs{pxs}} for the remainder of the paper.
When a measurement \ensuremath{\acs{scantupledef}} is received, the endpoint is estimated and used as an index to the distance map.
Assuming a Gaussian error distribution, the weight of each particle \ensuremath{\acs{prtnowprime}} can then estimated as
\vspace{-0.4cm}
\begin{equation}
 \acs{probrng} = \myexp{\slfrac{{-\dist{\mysmallscript{\occ}}    {2}}}{{2\sig{\mysmallscript{\occ}    }{2}}}}
 \label{eq:meth:probrng}
 \vspace{-0.15cm}
\end{equation}
where \ensuremath{\acs{distocc}} is the value obtained from the distance map and \ensuremath{\acs{sighit}} is dictated by the noise characteristics of the sensor.
However, this model has three main limitations.
First, it makes no use of the semantic information present in the map.
Second, the parameter \ensuremath{\acs{sighit}} must be estimated by the user and assumes all measurements within a scan have the same noise parameters.
Third, it is incapable of operating in the absence of range measurements.

Instead, this work uses the semantic labels present in the map to create multiple likelihood fields.
For each label present in the floorplan, we can calculate a distance map that stores the shortest distance to a cell with the same label.
Formally, for each map cell \ensuremath{\acs{cellat}} we can estimate the distance to the nearest cell of each label as 
	\vspace{-0.2cm}
\begin{equation}
	\acs{distlbl}\mypar{\acs{pxs}} = \min_{\pxs{}{}{}{\prime}}\mymag{\acs{pxs}-\acs{pxsprime}}, \quad \acs{cellatlblprime} > \thr{o}
	\label{eq:meth:distlbl}
	\vspace{-0.2cm}
\end{equation}
where \ensuremath{\acs{distlbl}= \mybra{\acs{distwll}, \acs{distdor}, \acs{distwdw}}} are distances to the nearest wall, door and window, respectively.
Figure \ref{fig:meth:floorplans} shows the distance maps for each label.
This approach overcomes the three limitations of the state-of-the-art, which we will now discuss.

\subsubsection{Semantic Information}
First, \ac{sedar} uses the semantic information present in the map.
When we receive an observation \ensuremath{\acs{semtupledef}}, we use the bearing \ensuremath{\acs{brnnowidx}} and range \ensuremath{\acs{rngnowidx}} information to estimate the endpoint of the scan.
We then use the label \ensuremath{\acs{lblnowidx}} to decide which semantic likelihood field to use.
Using the endpoint from the previous step, the label-likelihood can be estimated similarly to equation \ref{eq:meth:probrng},
	\vspace{-0.1cm}
\begin{equation}
 \acs{problbl} = \myexp{\slfrac{{-\dist{\mysmallscript{\lbl{}{}}}{2}}}{{2\sig{\mysmallscript{\lbl{}{}}}{2}}}}
 \label{eq:meth:prlbl}
	\vspace{-0.1cm}
\end{equation}
where \ensuremath{\acs{distlbl}} is the distance to the nearest cell of the relevant label and \ensuremath{\acs{siglbl}} is the standard deviation (which we will define using the label prior).
The probability of an observation given the map and pose can then be estimated as
	\vspace{-0.1cm}
\begin{equation}
  \!\acs{probray} \!\!=\! \acs{whtrng}\acs{probrng} \!\!+\! \acs{whtlbl}\acs{problbl}\!\!\!\!
	\vspace{-0.1cm}
\end{equation}
where \ensuremath{\acs{whtrng}} and \ensuremath{\acs{whtlbl}} are user defined weights.
When \ensuremath{\acs{whtlbl} = 0} the likelihood is the same as standard \ac{rmcl}.
On the other hand, when \ensuremath{\acs{whtrng} = 0} the approach is using only the semantic information present in the floorplan.
These weights are properly explored and defined in section \ref{sec:res:glo}.
Unlike range scanners, \ensuremath{\acs{siglbl}} cannot be related to the physical properties of the sensor.
Instead, this standard deviation is estimated directly from the prior of each label on the map.
Defining \ensuremath{\acs{siglbl}} this way has the benefit of not requiring tuning.
However, there is a much more important effect that must be discussed.
\subsubsection{Semantically Adaptive Standard Deviation}
When a human reads a floorplan, unique landmarks are the most discriminative features: it is easier to localise on a floorplan from the configuration of doors and windows than it is from the configuration of walls.
This translates into the a simple insight: \textit{lower priors are more discriminative}.
Therefore, \ensuremath{\acs{siglbl}} is tied to the prior of each label not only because it is one less parameter to tune, but because it implicitly makes observing rare landmarks more beneficial than common landmarks.

Relating \ensuremath{\acs{siglbl}} to the label prior \ensuremath{\acs{priorlbl}} controls how smoothly the distribution decays \wrt distance from the cell.
The smaller \ensuremath{\acs{priorlbl}} is, the smoother the decay.
In essence, the localisation algorithm should be more \textit{lenient} on sparser labels.\looseness=-1
	\vspace{-0.1cm}
\subsection{Range-less Semantic Scan-Matching}\label{sec:meth:mcl:sm:lbl}
	\vspace{-0.1cm}
The final, and most important, strength of this approach is the ability to perform all of the previously described methodology in the complete absence of range measurements.
So far, we have formalised this approach on the assumption that we received either \ensuremath{\acs{scantuple}} tuples (existing approaches) or \ensuremath{\acs{semtuple}} tuples (\ac{sedar}-based approach).
However, this approach is capable of operating directly on \ensuremath{\acs{semtuplenorange}} tuples.
In other words, depth measurments are \emph{explicitly} not added to this approach.

Incorporating range-less measurements is simple.
The beam and likelihood field models are combined in a novel approach that avoids the degeneracies that would happen in traditional \ac{rmcl} approaches.
In the standard approach, the raycasting operation terminates when an occupied cell is reached and the likelihood is estimated as
	\vspace{-0.3cm}
\begin{equation}
 \acs{probray} = \acs{zhit}
	\vspace{-0.1cm}
\end{equation}
where \ensuremath{\acs{rngnowidx}} is the range obtained from the sensor and \ensuremath{\acs{rngnowidxstar}} is the distance travelled by the ray.
Unfortunately, in the absence of a range-based measurement \ensuremath{\acs{rngnowidx}} this is impossible.
Using the standard distance map is also impossible, since we can not estimate the endpoint of the ray.
Using raycasting in the distance map fails similarly. 
The raycasting terminates on an occupied cell, implying \ensuremath{\acs{distocc} = 0} for every ray cast.

On the other hand, the semantic likelihood fields can still be used as \ensuremath{\acs{distlbl}} will still have a meaningful and discriminative value.
We call this operation semantic raycasting.
For every \ensuremath{\acs{semtuplenorangedef}}, the raycasting is performed.
However, instead of comparing \ensuremath{\acs{rngnowidx}} and \ensuremath{\acs{rngnowidxstar}} or using \ensuremath{\acs{distocc}}, the label \ensuremath{\acs{lblnowidx}} determines which likelihood field to use.
The cost is then
	\vspace{-0.1cm}
\begin{equation}
 \acs{probray} = \acs{problbl}
 \label{eq:sec:meth:mcl:probray}
	\vspace{-0.1cm}
\end{equation}
where \ensuremath{\acs{problbl}} is defined in equation \ref{eq:meth:prlbl}.
This method is essentially a combination of the beam-model and the likelihood field model.
In the absence of range-measurements to estimate an endpoint from, this hybrid approach uses semantic raycasting to find the nearest occupied cell.
The distances are then used to provide smoothness to equation \ref{eq:sec:meth:mcl:probray}, which implies that the observation likelihood is directly proportional to the angular distribution of labels.
The net effect is that this approach is invariant to scale changes, as long as the aspect ratio of the map is respected.\looseness=-1

To summarise, this section presented several important concepts.
We introduced the idea of a semantic floorplan that contains information that is salient to humans.
We also introduced a new sensing modality, \ac{sedar}, that adds semantic labels to the traditional \ac{lidar} information.
We then incorporated these two ideas into a novel \ac{mcl}-based approach.
This approach is capable of using the semantic information present in the map to define a novel motion model.
It is also capable of using the labels from a \ac{cnn}-based segmentation to localise within the map.
Our approach can do all of the above both in the presence, and absence, of range measurements.
In the following section, we show that our approach is capable of outperforming standard \ac{rmcl} approaches when using depth, and that it provides comparable performance in its absence.

\vspace{-0.2cm}
\section{Results}\label{sec:res}
	\vspace{-0.1cm}
This section will demonstrate that \ac{sedar}-based \ac{mcl} is capable of reliably out-performing the state-of-the-art when using range measurements.
It will also show that our approach it is capable of comparable performance even in the absence of range.
First, the experimental setup is described.
This consists of creating a dataset of a trajectory within a floorplan, as well as establishing error metrics.
Then a comparison of several approaches is performed. 
The comparison is done in terms of room-level and global localisation, both quantitative and qualitative.
Finally, we show the effects of our parameters.

	\vspace{-0.1cm}
\subsection{Experimental Setup}
	\vspace{-0.1cm}
In order to evaluate this approach, we require a dataset that has several important characteristics.
The dataset should consist of a robot navigating within a human-readable floorplan.
Human-readability is required to ensure semantic information is present.%
The trajectory should be captured with an RGB-D camera in order to extract all the possible tuple combinations (range, bearing and label).
Finally, we expect the trajectory of the robot to happen on the same plane as the floorplan.
Unfortunately, most of the \ac{mcl} datasets in the literature do not contain a floorplan, opting instead for laser-scans.
RGB-D \ac{slam} datasets are more appropriate, but they either do not move on the floorplan plane or simply do not contain ground-truth trajectory estimation.

Therefore, we are forced to use our own dataset - which we will make publicly available.
We use the floorplan in figure \ref{sec:meth:fig:floorplan} because it is large enough to provide multiple trajectories with no overlap.
The dataset was collected using the popular TurtleBot platform, as it has a front-facing Kinect that can be used for emulating both \ac{lidar} and \ac{sedar}.

Normally, the ground-truth trajectory for floorplan localisation is either manually estimated (as in \cite{Winterhalter2015}) or estimated using \ac{mocap} systems (as in \cite{Sturm2012}).
However, both of these approaches are limited in scope.
Manual ground-truth estimation is time-consuming and impractical.
\ac{mocap} is expensive, difficult to calibrate, and normally cannot remain in the public areas required for floorplan localisation.
In order to overcome these limitations, well established RGB-D \ac{slam} systems are used instead.
The excellent approach by Labbe \etal \cite{labbe2014} provides very accurate pose estimation in complex environments.
While it does not localise within a floorplan, it does provide an accurate reconstruction and trajectory for the robot, which can then be registered into the floorplan.

To quantitatively evaluate the presented approach against ground truth, the \ac{ate} metric presented by Sturm \etal \cite{Sturm2012} is used.
\ac{ate} is estimated by first registering the two trajectories using the closed form solution of Horn \cite{Horn1987}, who finds a rigid transformation \ensuremath{\acs{posngttf}} that registers the trajectory \ensuremath{\acs{posnall}} to the ground truth \ensuremath{\acs{posngtall}}.
At every time step \ensuremath{\mytime{}}, the \ac{ate} can then be estimated as
	\vspace{-0.3cm}
\begin{equation}
 \sca{}{}{e}{\posngt{}{}}{} = \posngt{\mytime}{\hspace{-1mm}\inverse} \acs{posngttf} \acs{posnnow}
	\vspace{-0.2cm}
\end{equation}
where \ensuremath{\acs{posngtnow} \in \acs{posngtall}} and \ensuremath{\acs{posnnow} \in \acs{posnall}} are the current time-aligned poses of the ground truth and estimated trajectory, respectively.
The \ac{rmse}, mean and median values of this error metric are reported, as these are indicative of performance over room-level initialisation.
In order to visualise the global localisation process, the error of each successive pose is shown (error as it varies with time).
These metrics are sufficient to objectively demonstrate the systems ability to globally localise in a floorplan, while also being able to measure room-level initialisation performance.

We compare the work presented here against the extremely popular \ac{mcl} approach present in \ac{ros}, called \ac{amcl} \cite{Dellaert1999}.
While more modern approaches \cite{Blanco2008} exist, they are based on the same principles as \ac{amcl} and simply change the particle sampling strategy.
More importantly, \ac{amcl} is the standard \ac{mcl} approach in the robotics community. 
Any improvements over this approach are therefore extremely valuable.
In all experiments, any overlapping parameters (such as \ensuremath{\acs{sighit}}) are kept the same.
The only parameters varied are \ensuremath{\acs{whtlbl}}, \ensuremath{\acs{whtrng}} and \ensuremath{\acs{ghtfac}}.

\vspace{-0.3cm}
\subsection{Room-Level Initialisation}\label{sec:res:cor}
\vspace{-0.2cm}
\begin{table}
\resizebox{\linewidth}{!}{
\centering
\begin{tabular}{|c|c|c|c|c|c|c|}
\hline
\multicolumn{7}{|c|}{\textbf{Average Trajectory Error (m)}}                                                                                      \\ \hline
\textbf{Approach}       		& \textbf{\acs{rmse}}	& \textbf{Mean}     & \textbf{Median}   & \textbf{Std. Dev.} & \textbf{Min}      & \textbf{Max}      \\ \hline
AMCL                    		& 0.24          	& 0.21          & 0.20          & 0.11           & 0.04          & 0.95          \\ \hline
\textbf{Range} (Label Only)      	& \textbf{0.19} 	& \textbf{0.16} & \textbf{0.14} & \textbf{0.10}  & \textbf{0.02} & \textbf{0.55} \\ \hline
\textbf{Range} (Combined)        	& 0.22          	& 0.19          & 0.17          & 0.11           & 0.04          & 0.62          \\ \hline
\textbf{Rays} \ensuremath{(\acs{ghtfac} = 3.0)} 	& 0.40          	& 0.34          & 0.27          & 0.22           & 0.07          & 1.51          \\ \hline
\textbf{Rays} \ensuremath{(\acs{ghtfac} = 7.0)} 	& 0.58          	& 0.45          & 0.38          & 0.37           & 0.02          & 2.23          \\ \hline
\end{tabular}
}
	\vspace{-0.15cm}
\caption{Room-Level Initialisation}
\label{sec:res:tbl:local}
	\vspace{-0.9cm}
\end{table}
For this evaluation, a room-level initialisation with standard deviations of \ensuremath{2.0 m} in \ensuremath{(x,y)} and \ensuremath{2.0 rad} in \ensuremath{\theta} is given to both \ac{amcl} and the proposed approach.
The systems then ran with a maximum of \ensuremath{1000} particles (minimum \ensuremath{250}) placed around the covariance ellipse.
We record the error as each new image in the dataset is added.
\begin{figure}[bt]
	\begin{center}
  \subfloat[Room-level]{\includegraphics[width=0.49\linewidth]{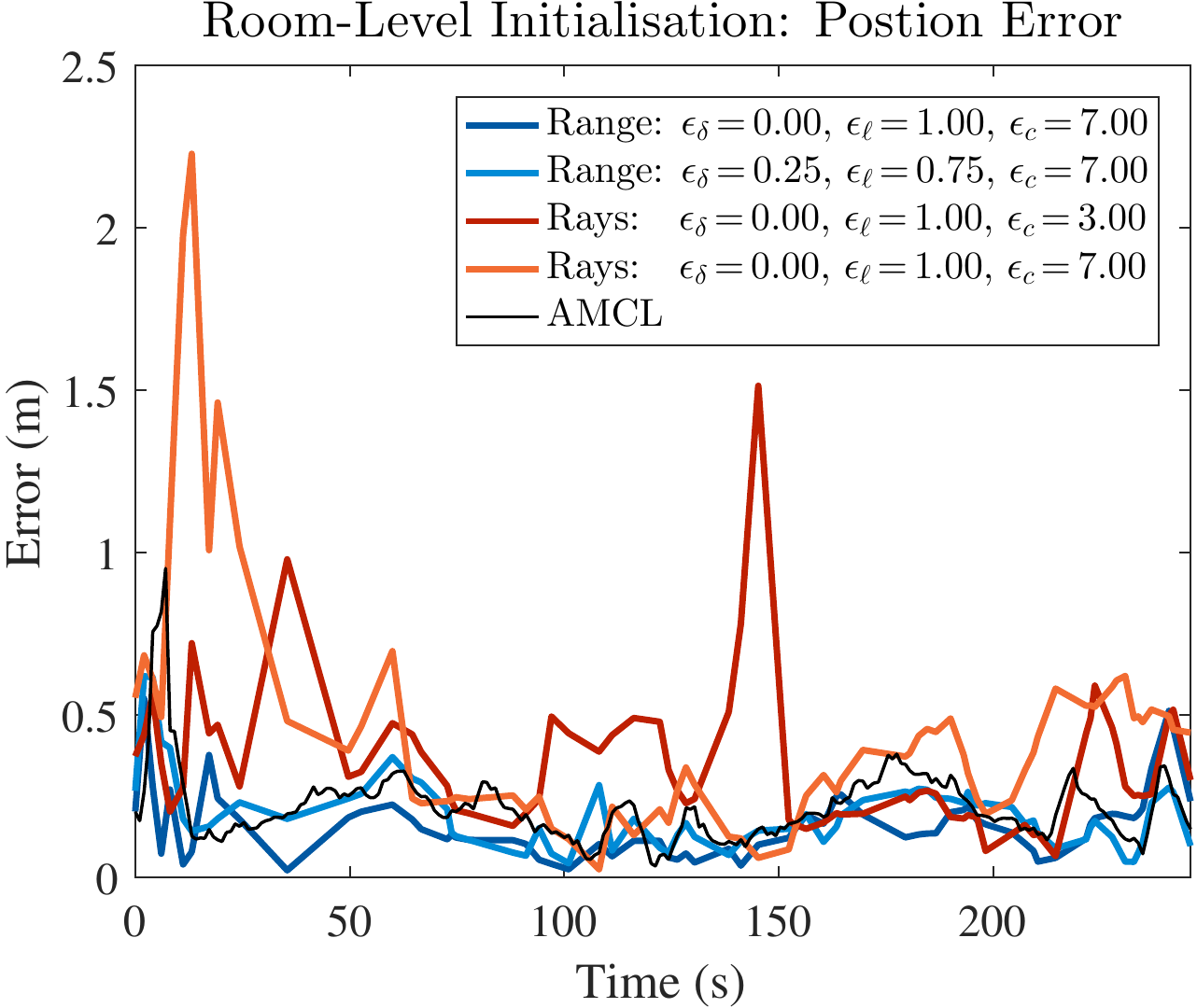}\label{sec:res:fig:local}}
  \subfloat[Global]{\includegraphics[width=0.49\linewidth]{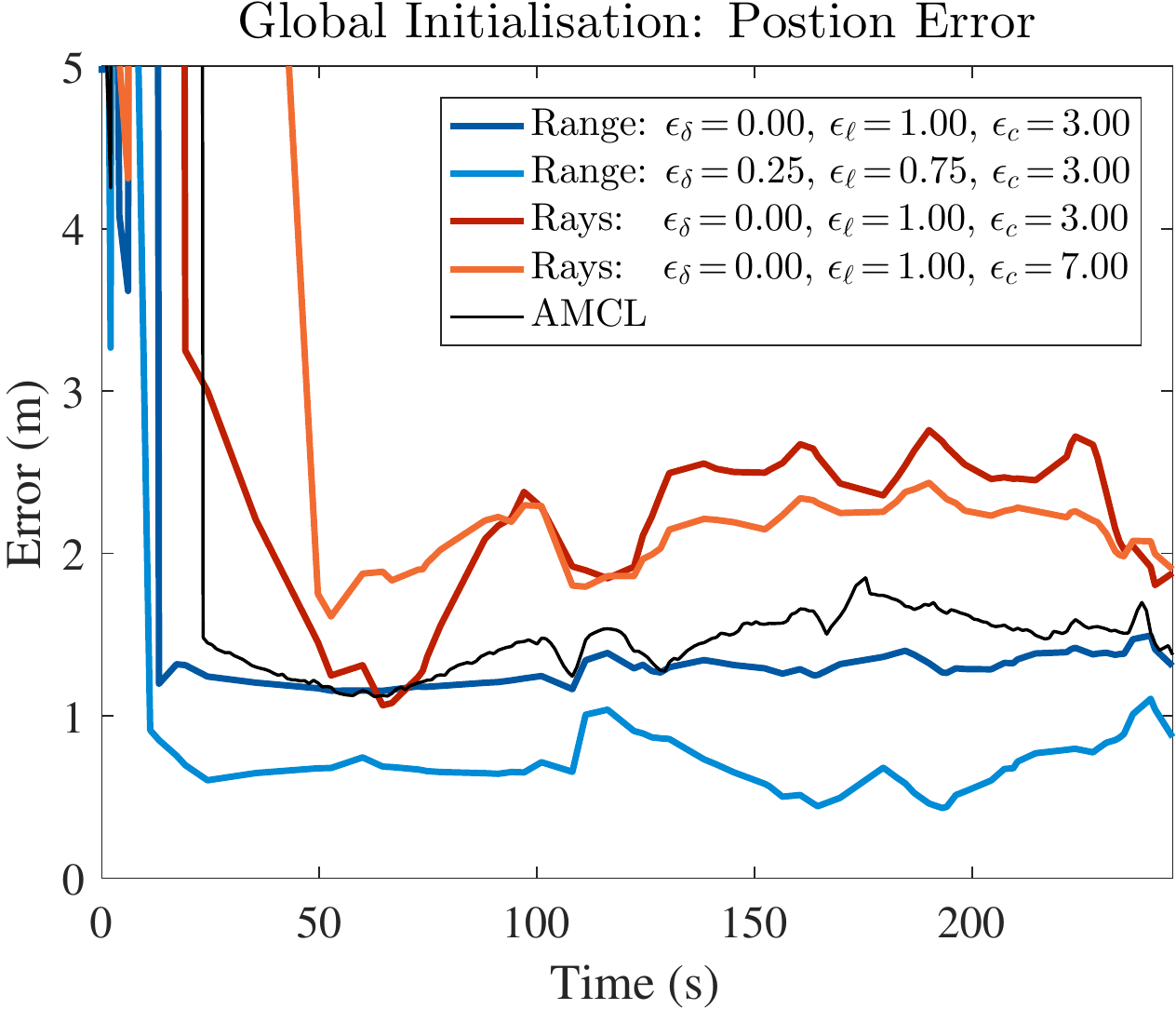}\label{sec:res:fig:global}}
	\end{center}
	\vspace{-0.35cm}
  \caption{Semantic localisation with different initialisations.}  
  \label{sec:res:fig:both}
	\vspace{-0.6cm}
\end{figure}

\subsubsection{Quantitative Results}
Figure \ref{sec:res:fig:local} compares four distinct scenarios against \ac{amcl}.
Of these four scenarios, two use the range measurements from the Microsoft Kinect (blue lines) and two only use the RGB image (red lines). 

The first range-enabled scenario uses the range measurements to estimate the endpoint of the measurement (and therefore the index in the distance map) and then sets \ensuremath{(\acs{whtrng} = 0.0, \acs{whtlbl} = 1.0)}. 
This means that while the range information is used to inform the lookup in the distance map, the costs are always directly related to the labels. 
The second range-enabled scenario performs a weighted combination \ensuremath{(\acs{whtrng}=0.25, \acs{whtlbl}=0.75)} of both the semantic and traditional approaches.\looseness=-1

In terms of the ray-based version of our approach, we use equation \ref{eq:sec:meth:mcl:probray}.
This means there are no parameters to set. 
Instead, a mild ghost factor (\ensuremath{\acs{ghtfac} = 3.0}) and a harsh one (\ensuremath{\acs{ghtfac} = 7.0}) are shown.

Since room-level initialisation is an easier problem than global initialisation, the advantages of the range-enabled version of our approach are harder to see compared to state-of-the-art.
However, it is important to notice how closely the ray-based version of the approach performs to the rest of the scenarios, despite using no depth data.
Apart from a couple of peaks, we essentially perform at the same level as \ac{amcl}.
This becomes even more noticeable in table \ref{sec:res:tbl:local}, where it is clear that range-based semantic \ac{mcl} (using only the labels) outperforms state of the art, while the ray-based \ensuremath{\acs{ghtfac} = 3.0} version lags closely behind.
The reason \ensuremath{\acs{ghtfac} = 3.0} performs better than \ensuremath{\acs{ghtfac} = 7.0} is because small errors in the pose can cause the robot to ``clip'' a wall as it goes through the door.
Since \ensuremath{\acs{ghtfac} = 3.0} is more lenient on these scenarios, it is able to outperform the harsher ghost factors.
We will explore this relationship further in section \ref{sec:res:ght}.
\begin{figure}
	\begin{center}
		\subfloat[Room-Level Initialisation]{\includegraphics[width=0.45\linewidth]{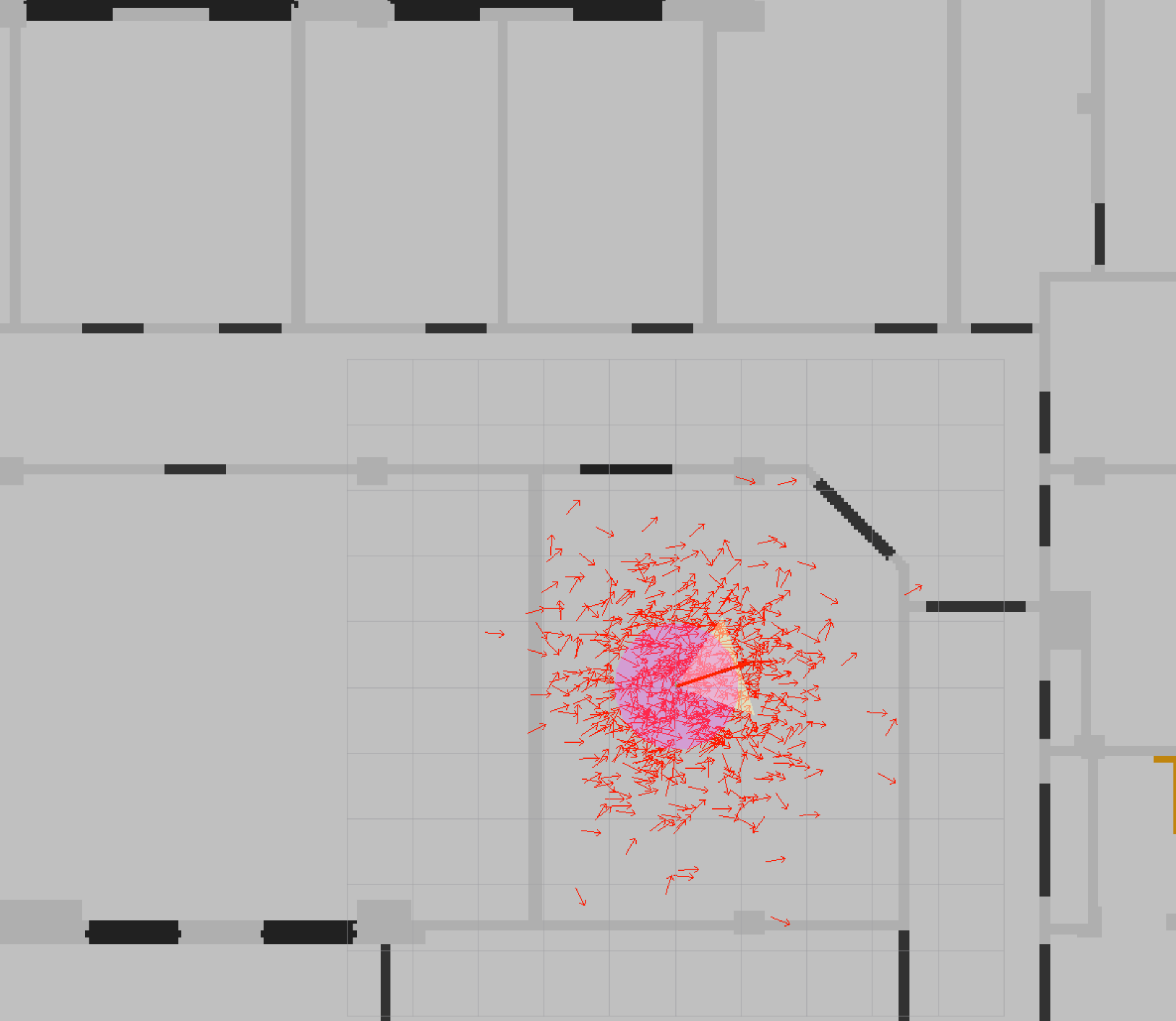}\label{sec:res:fig:rec_1}}\hspace{1mm}
		\subfloat[AMCL]{\includegraphics[width=0.45\linewidth]{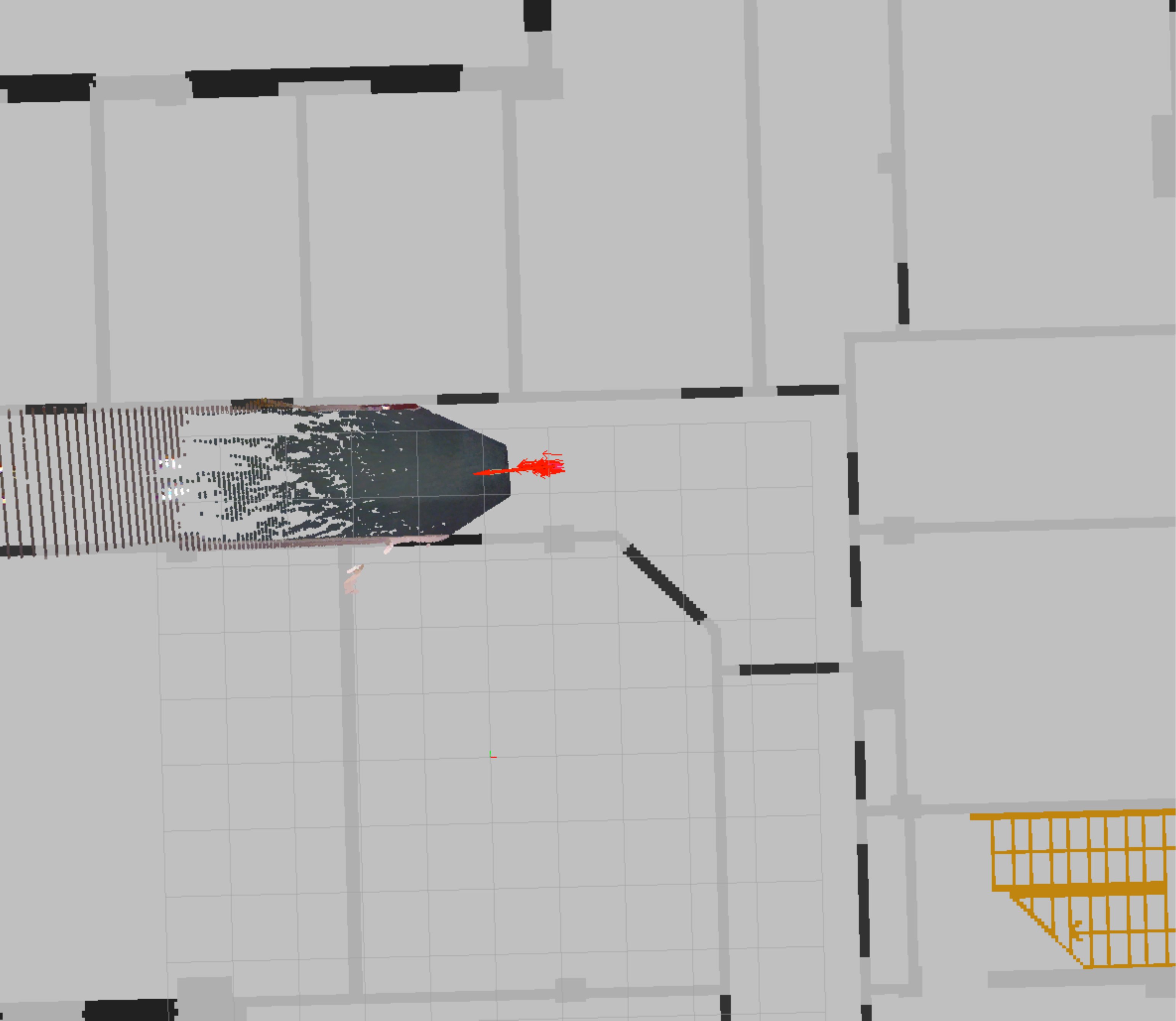}\label{sec:res:fig:rec_2}}\\
	\vspace{-0.3cm}
		\subfloat[\ac{sedar} (Range-Based)]{\includegraphics[width=0.45\linewidth]{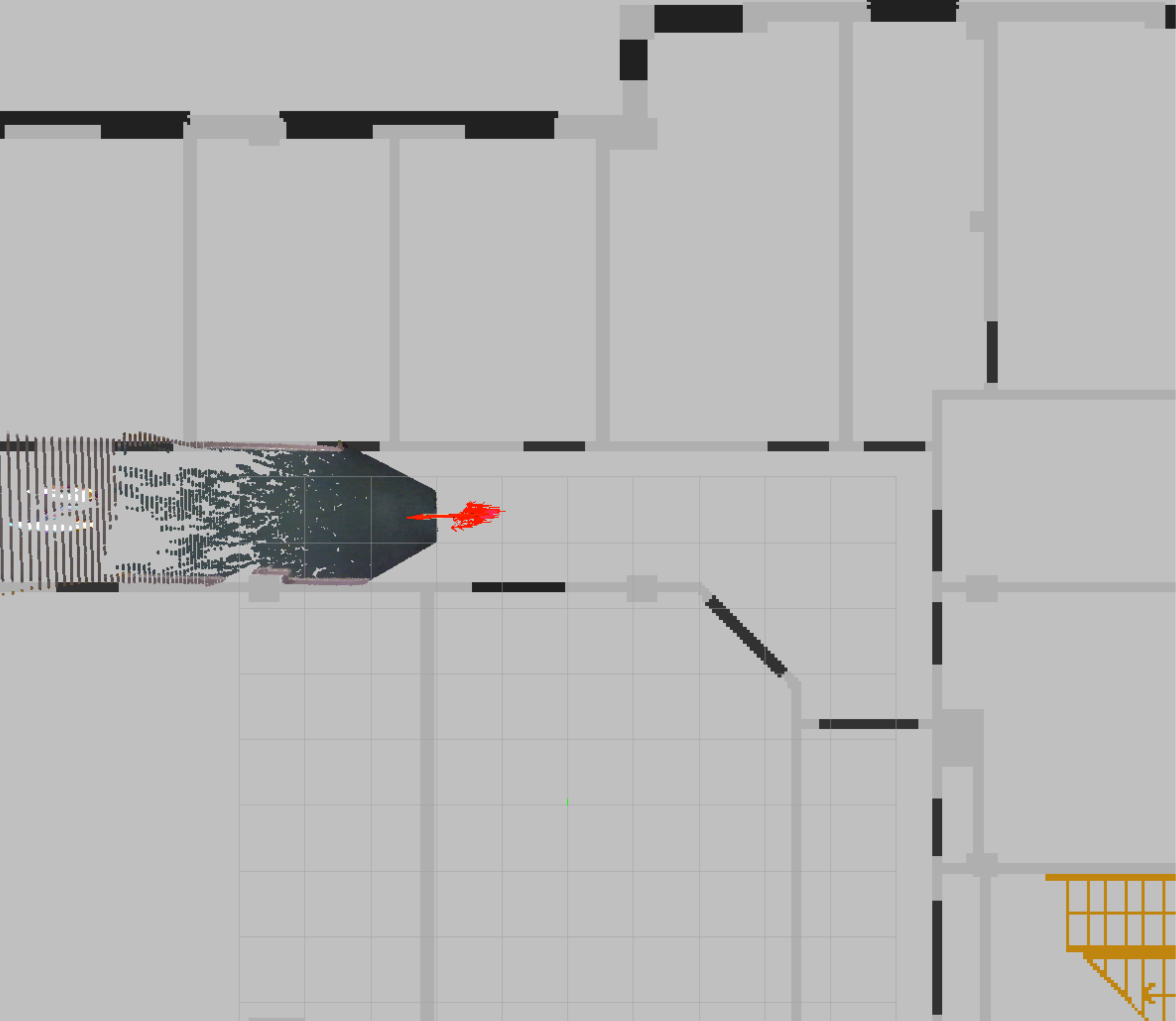}\label{sec:res:fig:rec_3}}\hspace{1mm}
		\subfloat[\ac{sedar} (Ray-Based)]{\includegraphics[width=0.45\linewidth]{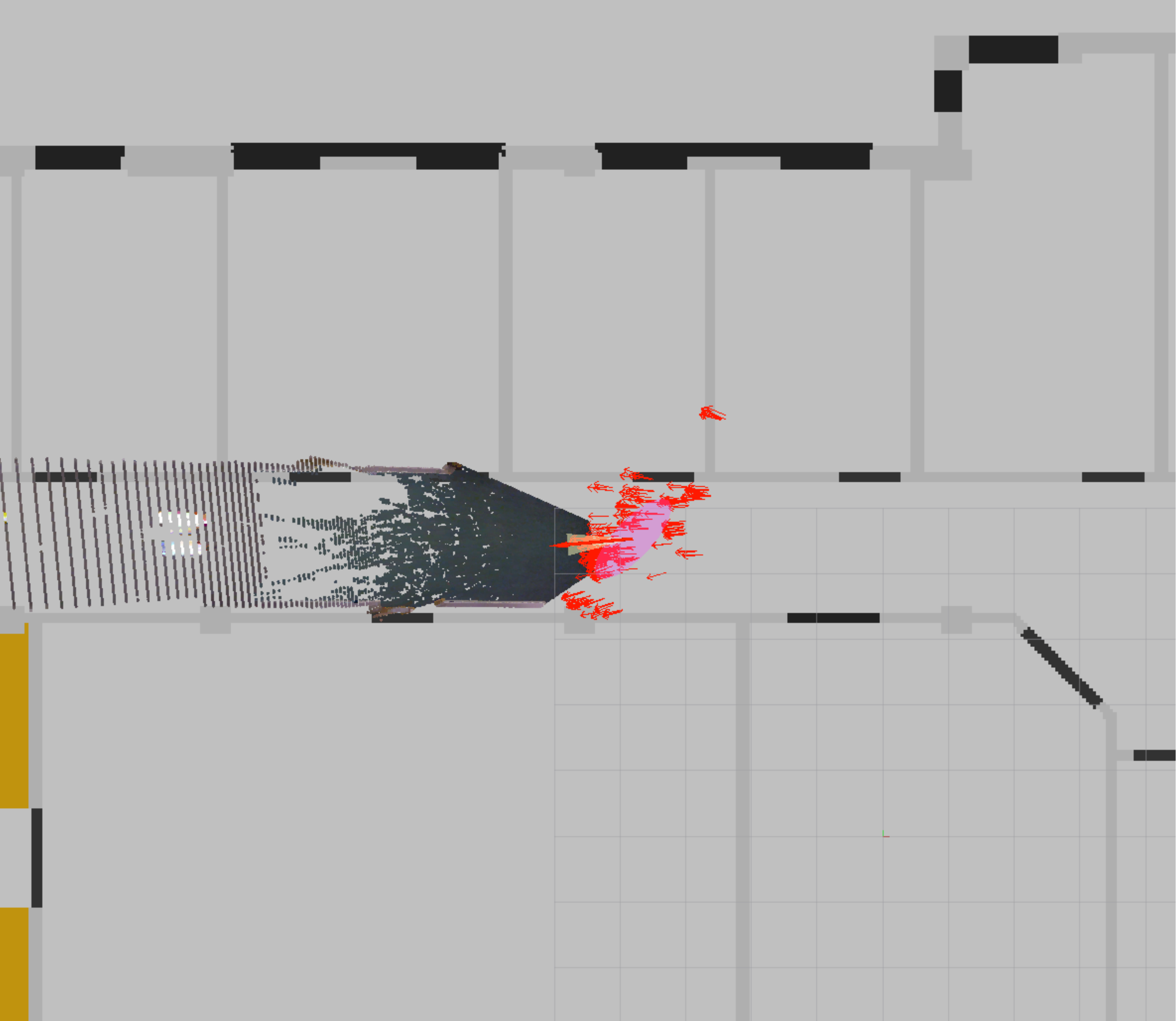}\label{sec:res:fig:rec_4}}
	\end{center}  
	\vspace{-0.3cm}
	\caption{Qualitative view of Localisation in different modalities.}  
	\vspace{-0.6cm}
	\label{sec:res:fig:rec}
\end{figure}
\begin{figure}
	\begin{center}
		\subfloat[AMCL]{\includegraphics[trim={10cm 11cm 0 0},clip,width=0.33\linewidth]{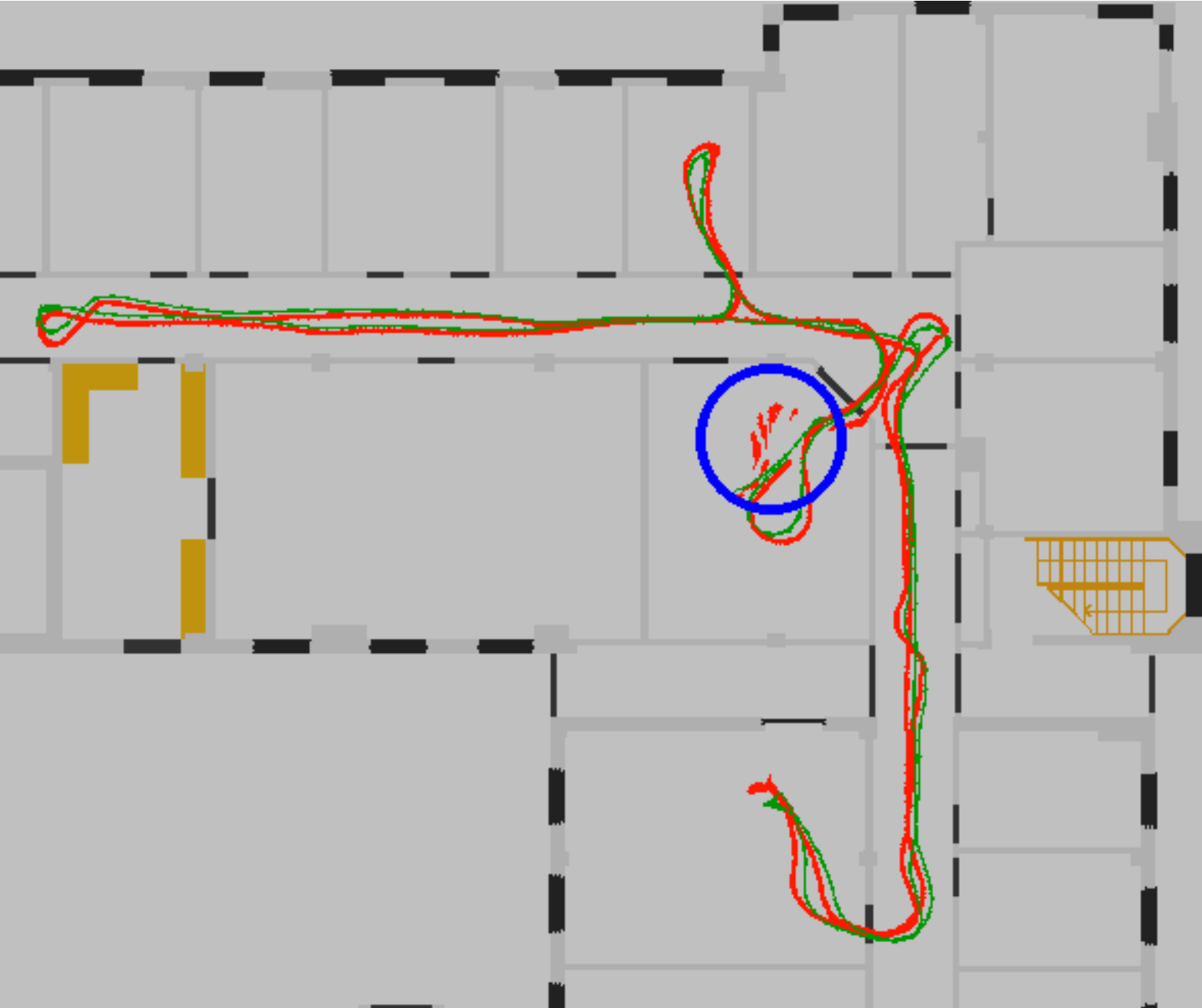}\hspace{1mm}\label{sec:res:fig:path:cor:amcl}}
		\subfloat[\ac{sedar} (Range)]{\includegraphics[trim={10cm 11cm 0 0},clip,width=0.33\linewidth]{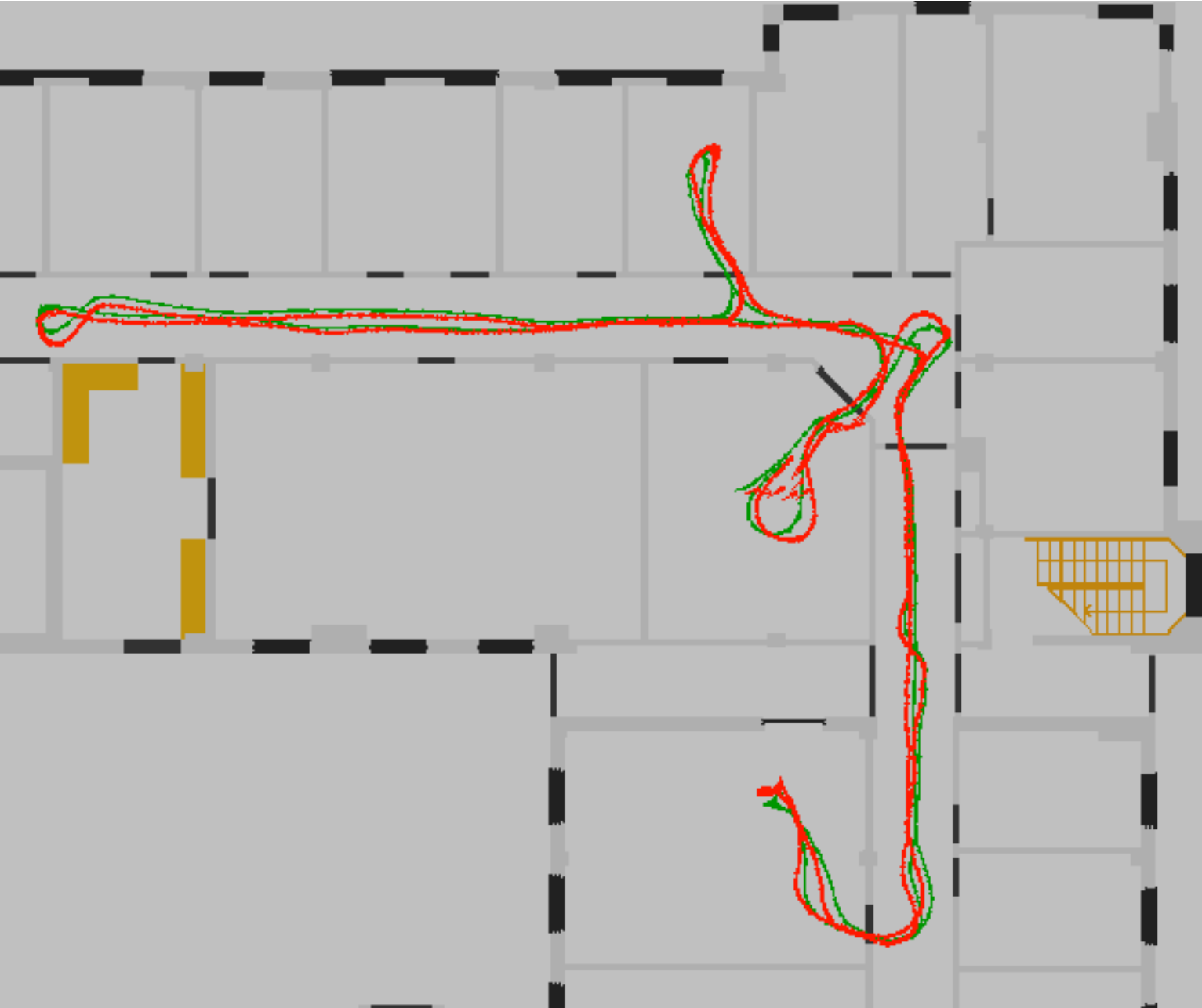}\hspace{1mm}\label{sec:res:fig:path:cor:mxd}}
		\subfloat[\ac{sedar} (Ray)]{\includegraphics[trim={10cm 11cm 0 0},clip,width=0.33\linewidth]{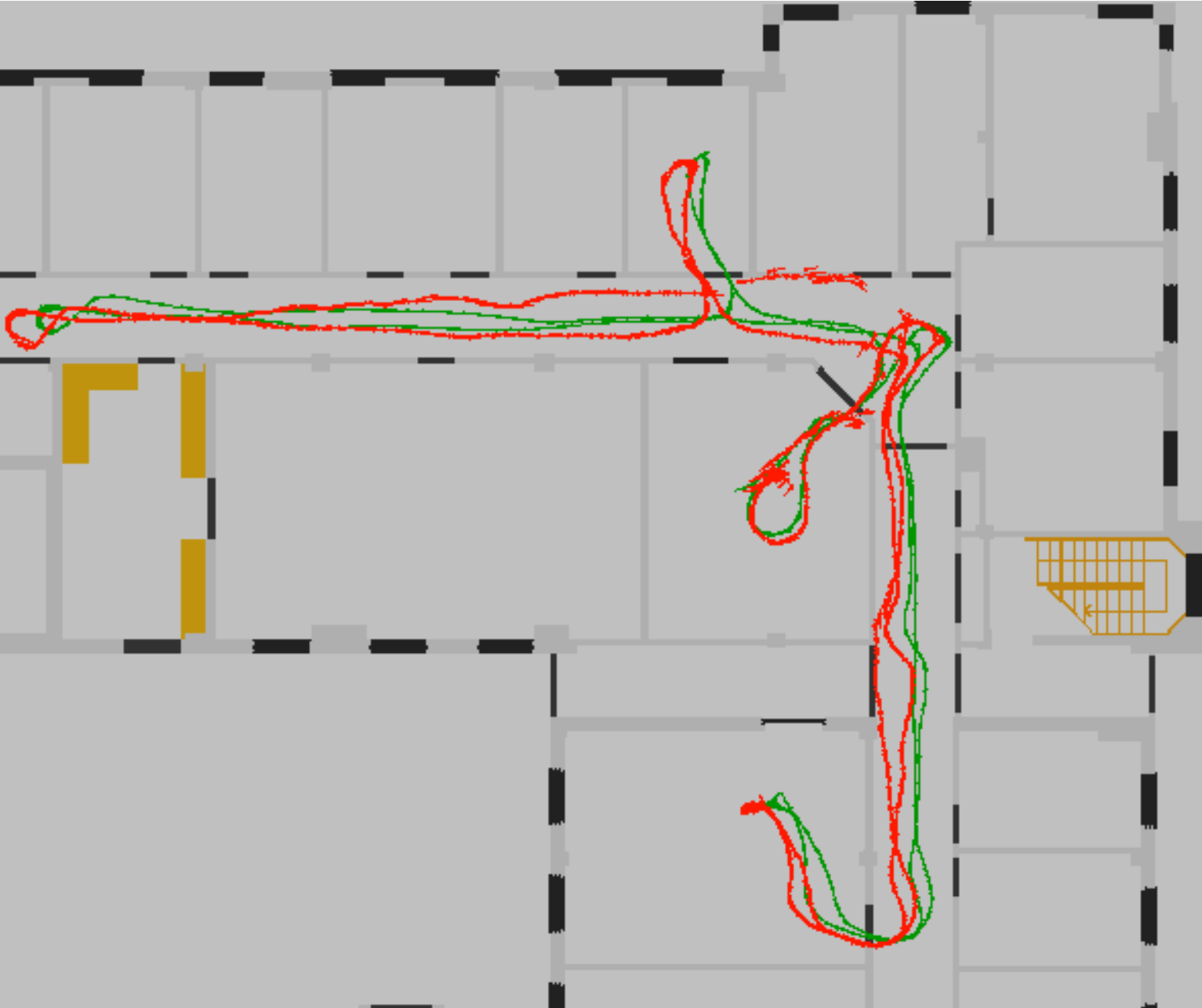}\label{sec:res:fig:path:cor:ray}}
	\end{center}  
	\vspace{-0.3cm}
	\caption{Estimated path from room-level initialisations.}  
	\vspace{-0.6cm}
	\label{sec:res:fig:path:cor}
\end{figure}

\subsubsection{Qualitative Results}
In terms of qualitative evaluation, we show the convergence behaviour and the estimated path.
The convergence behaviour can be seen in figure \ref{sec:res:fig:rec}.
Here, figure \ref{sec:res:fig:rec_1} shows how the filter is initialised to roughly correspond to the room the robot is in.
As the robot starts moving, we can see how \ac{amcl} (\ref{sec:res:fig:rec_2}), the range-based version of \ac{sedar} (\ref{sec:res:fig:rec_3}) and the ray-based version (\ref{sec:res:fig:rec_4}) converge.
Notice that while the ray-based approach has a predictably larger variance on the particles, the filter has successfully localised.
This can be seen from the fact that the reconstructed Kinect pointcloud is properly aligned with the floorplan.
It is important to note that although the Kinect pointcloud is present for visualisation in the ray-based method, it is \textit{not} used.\looseness=-1

The estimated paths can be seen in figure \ref{sec:res:fig:path:cor}, where the red path is the estimated path and green is the ground truth.
Figure \ref{sec:res:fig:path:cor:amcl} shows the state-of-the-art, which struggles to converge at the beginning of the sequence (marked by a blue circle).
It can be seen that the range-based approach in figure \ref{sec:res:fig:path:cor:mxd} (combined label and range), converges more quickly and maintains a similar performance to AMCL. 
It only slightly deviates from the path at the end of the ambiguous corridor on the left, which also happens to AMCL.
It can also be seen that the ray-based approach performs very well.
While it takes longer to converge, as can be seen by the estimated trajectory in figure \ref{sec:res:fig:path:cor:ray}, it corrects itself and only deviates from the path in areas of large uncertainty (like long corridors).

These experiments show that \ac{sedar}-based \ac{mcl} is capable of operating in a room-level initialised scenario. 
It is now important to discuss how discriminative \ac{sedar} is when there is no initial pose estimate provided to the system.

	\vspace{-0.2cm}
\subsection{Global Initialisation}\label{sec:res:glo}
	\vspace{-0.15cm}
We now focus on \ac{sedar}-based \ac{mcl}'s ability to perform global localisation.
In these experiments, the system is given no indication of where in the map the robot is.
Instead, a maximum $50k$ particles (min. $15k$) is placed over the floorplan.\looseness=-1
\subsubsection{Quantitative Results}
Figure \ref{sec:res:fig:global} shows the same four scenarios as in the previous section.
For the range-based scenarios (blue lines) it can be seen that using only the label information (\ensuremath{\acs{whtrng}=0.0, \acs{whtlbl}=1.00}) consistently outperforms the state of the art, both in terms of how quickly the values converge to a final result and the actual error on convergence.
This shows that \ac{sedar} used in an \ac{mcl} context is more discriminative than standard occupancy maps in \ac{rmcl}.
The second range-based measurement (\ensuremath{\acs{whtrng}=0.25, \acs{whtlbl}=0.75}) significantly outperforms all other approaches.
This is probably because, in principle, the occupancy maps can be considered another ``label'' in the semantic floorplan.
This makes sense because setting \ensuremath{\acs{whtrng}=0.25} is equivalent to weighting all labels equally, as it is a third of \ensuremath{\acs{whtlbl}=0.75} which is the weight of 3 labels.\looseness=-1
\begin{figure}
	\begin{center}
		\subfloat[Global Initialisation]{\includegraphics[width=0.32\linewidth]{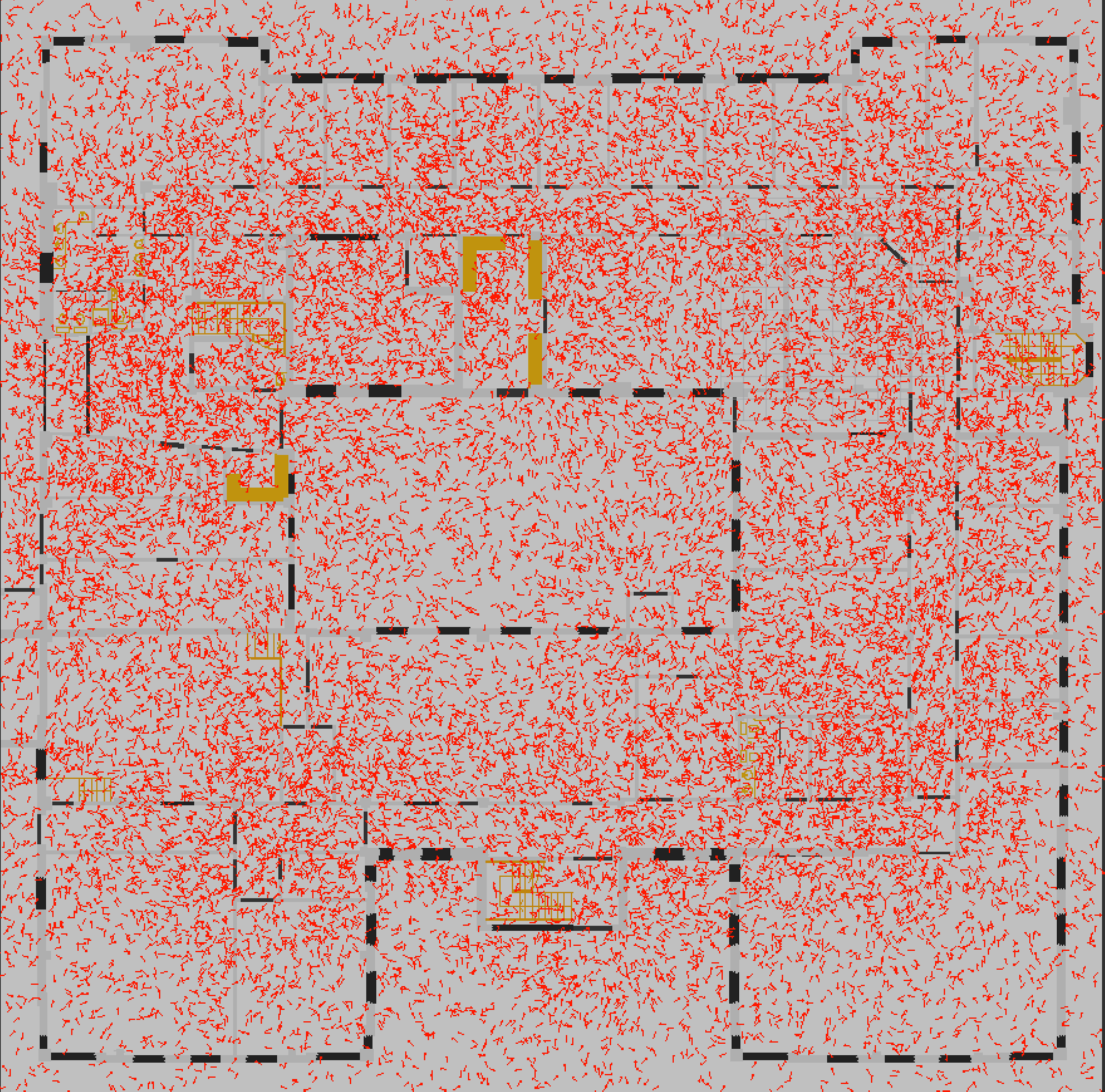}\label{sec:res:fig:conv_1}}\hspace{0.5mm}
		\subfloat[Looking at Doors]{\includegraphics[width=0.32\linewidth]{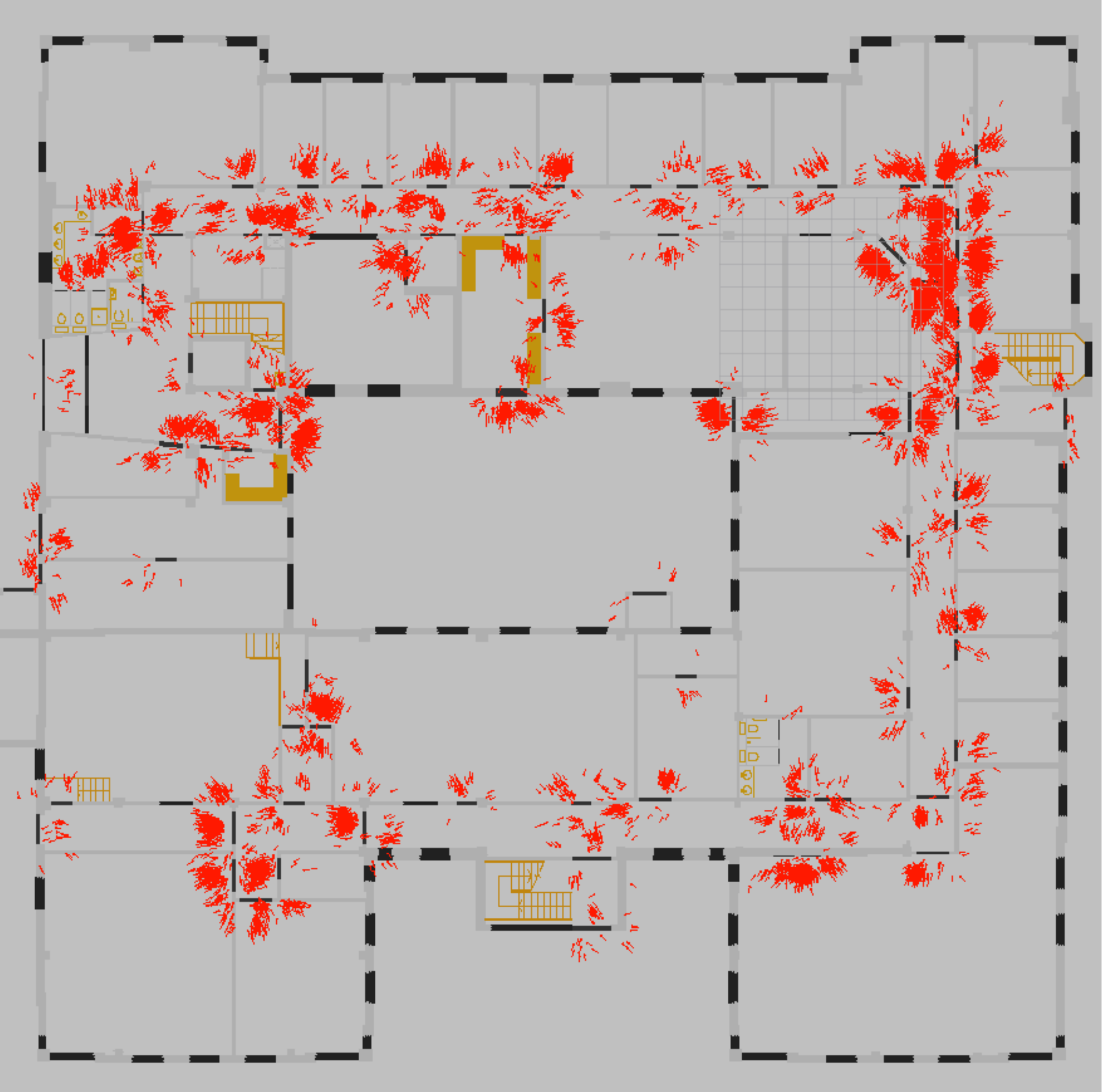}\label{sec:res:fig:conv_3}}\hspace{0.5mm}
		\subfloat[Converged]{\includegraphics[width=0.32\linewidth]{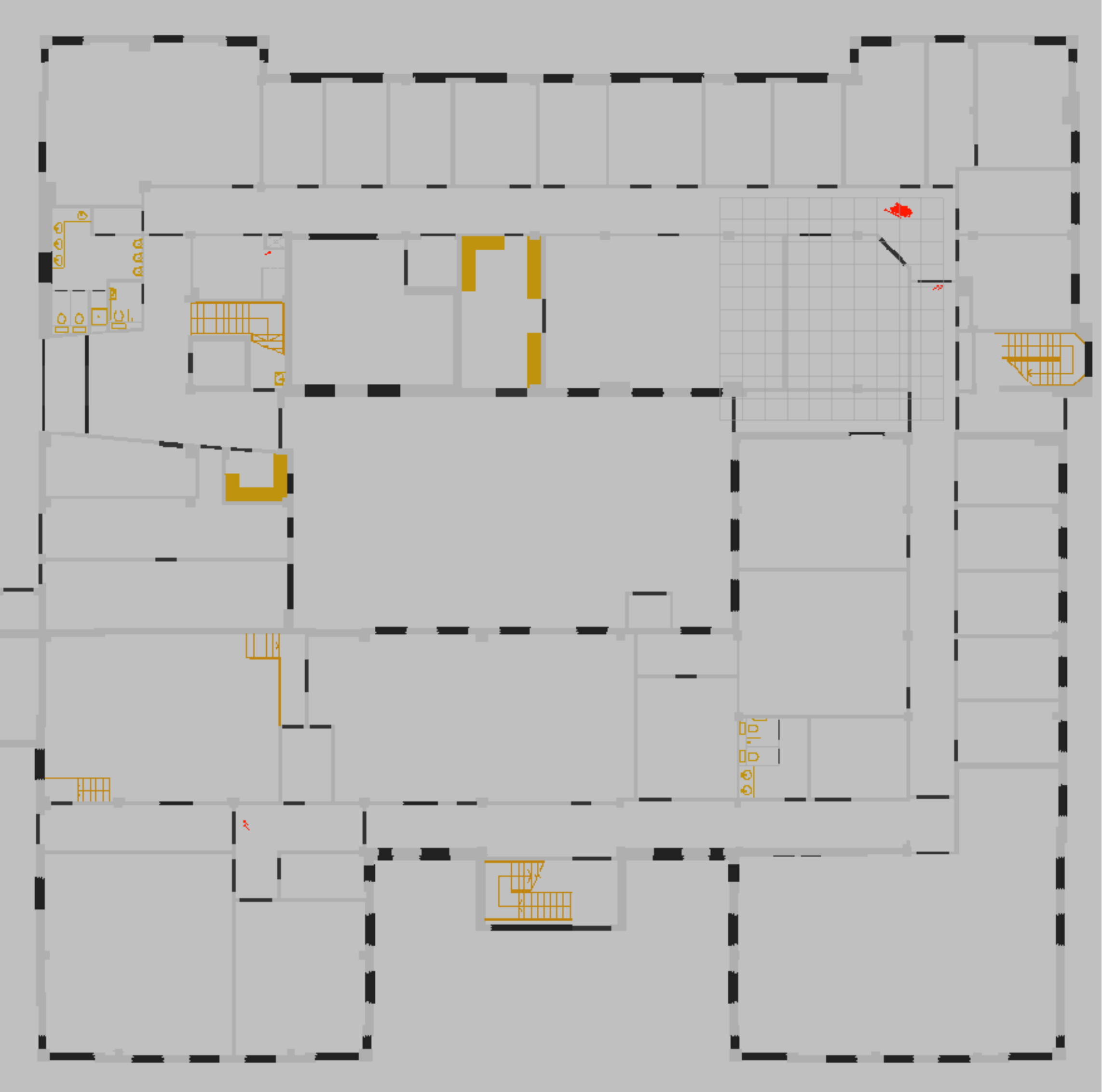}\label{sec:res:fig:conv_4}}
	\end{center}  
	\vspace{-0.3cm}
	\caption{Qualitative view of Localisation in different modalities.} 
	\vspace{-0.4cm} 
\end{figure}
\begin{figure}
	\begin{center}
		\subfloat[AMCL Path]{\includegraphics[trim={10cm 11cm 0 0},clip,width=0.32\linewidth]{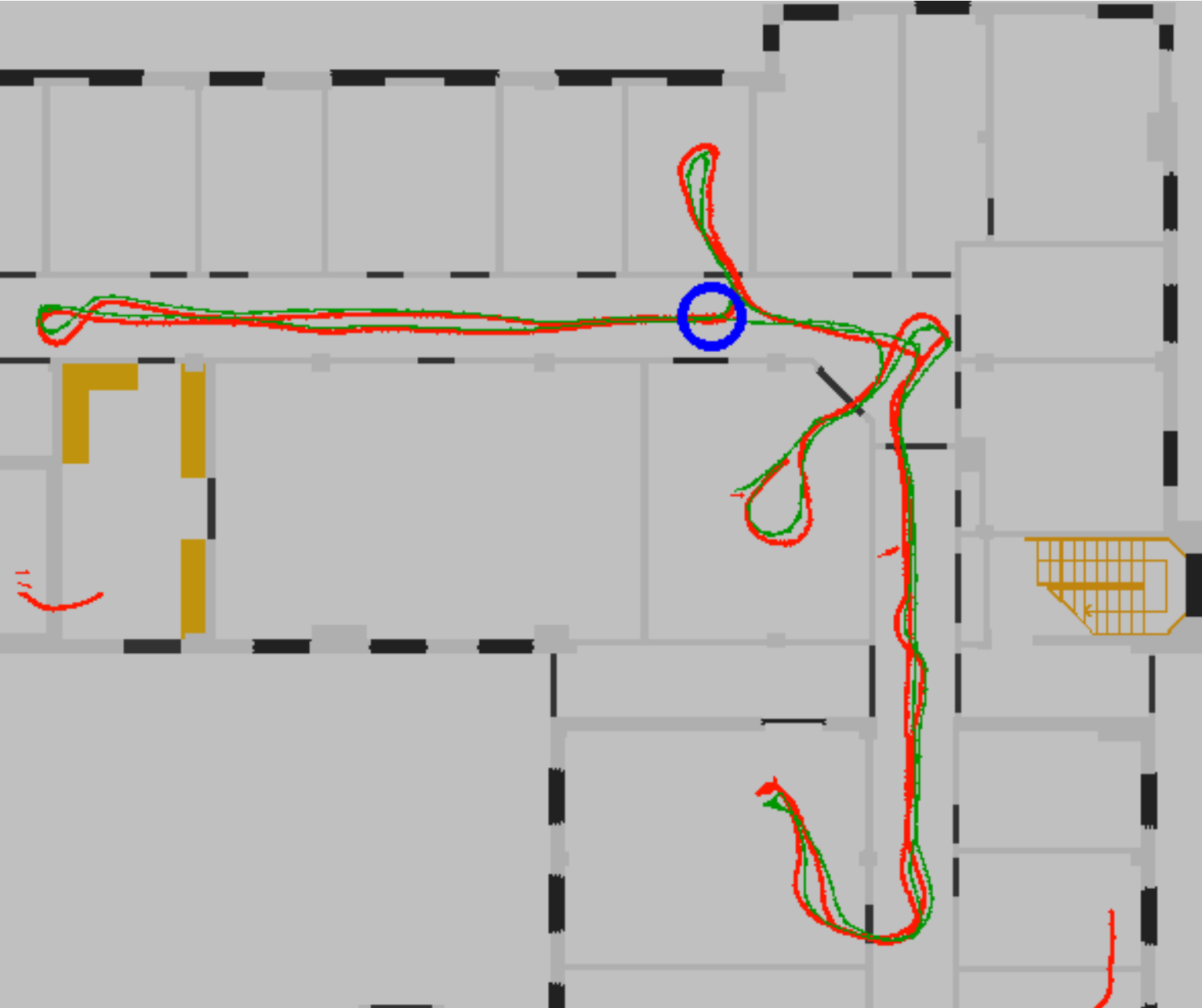}\label{sec:res:fig:path:gbl:amcl}}\hspace{0.5mm}
		\subfloat[\ac{sedar} (Range)]{\includegraphics[trim={10cm 11cm 0 0},clip,width=0.32\linewidth]{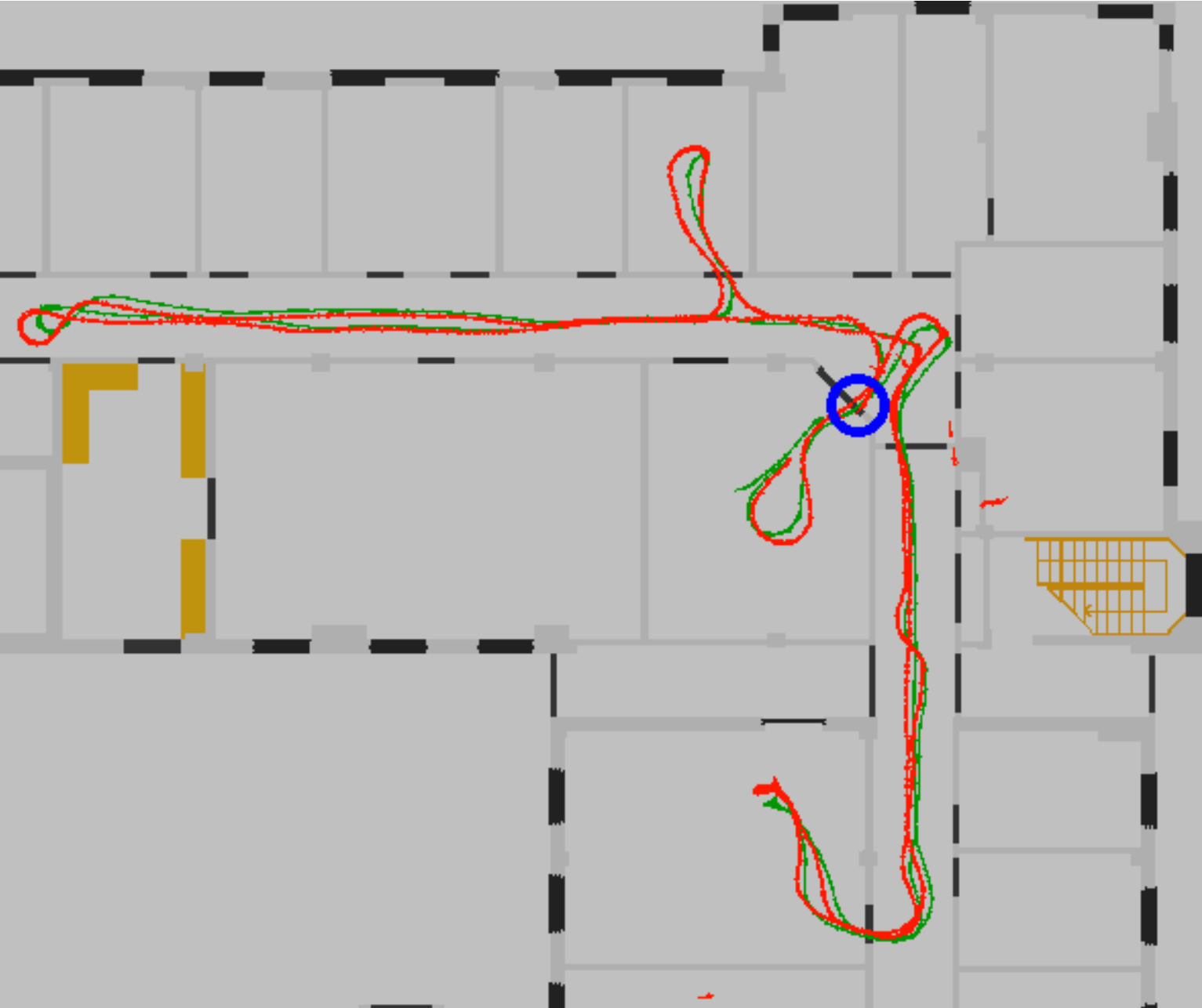}\label{sec:res:fig:path:gbl:mxd}}\hspace{0.5mm}
		\subfloat[\ac{sedar} (Ray)]{\includegraphics[trim={10cm 11cm 0 0},clip,width=0.32\linewidth]{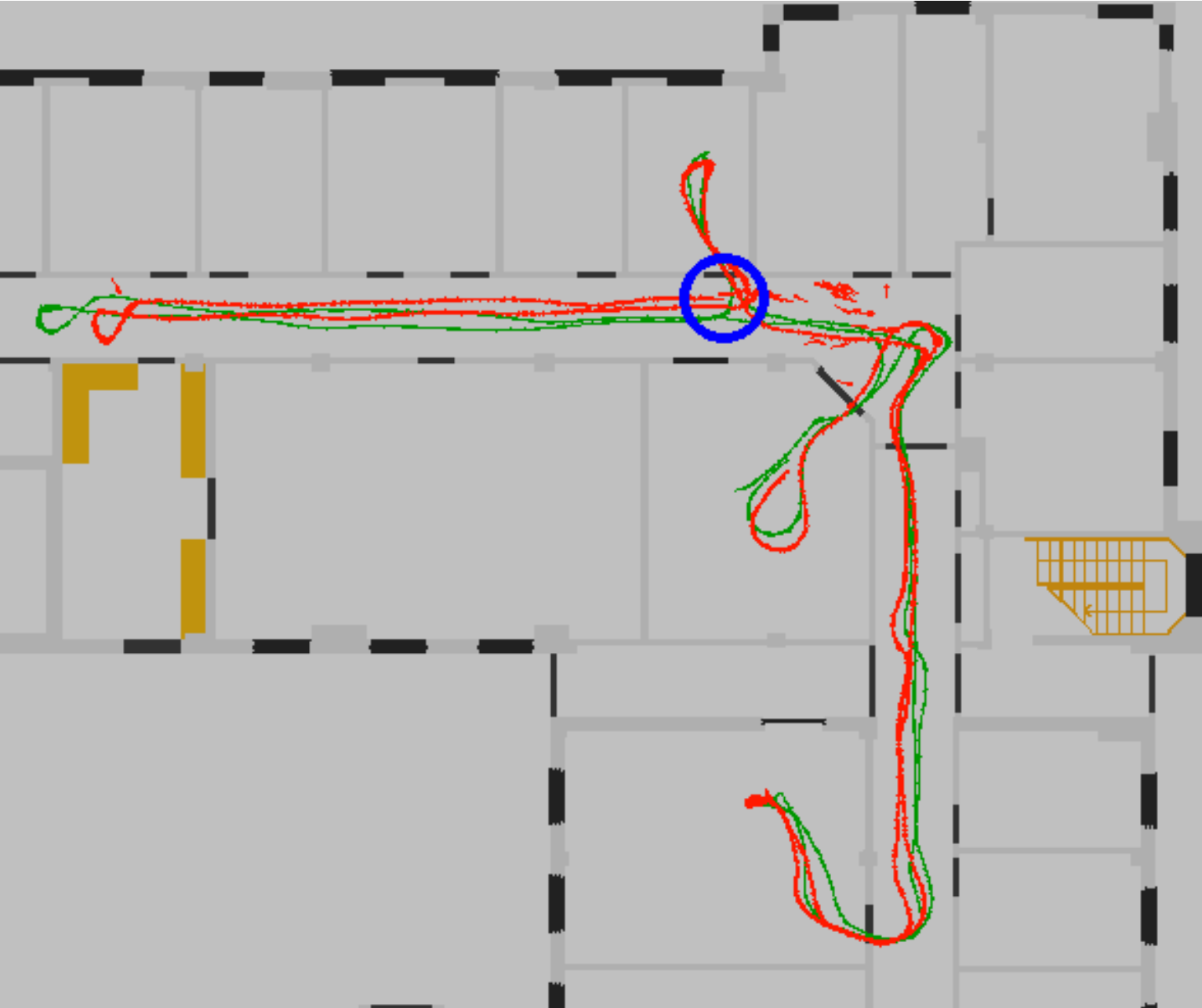}\label{sec:res:fig:path:gbl:ray}}
	\end{center}  
	\vspace{-0.3cm}
	\caption{Estimated path from global initialisations.}
	\vspace{-0.2cm}  
	\label{sec:res:fig:path:gbl}
\end{figure}
In terms of the ray-based version of our approach (red lines), we compare two scenarios.
A mild ghost factor (\ensuremath{\acs{ghtfac} = 3.0}) and a harsh one (\ensuremath{\acs{ghtfac} = 7.0}).
These versions of the approach both provide comparable performance to the state-of-the-art.
It is important to emphasise that this approach uses absolutely no range and/or depth measurements.
As such, comparing against depth-based systems is inherently unfair.
Still, \ac{sedar} ray-based approaches compare favourably to \ac{amcl}.
In terms of convergence, the mild ghost factor \ensuremath{\acs{ghtfac} = 3.0} gets to within several meters even quicker than \ac{amcl}, at which point the convergence rate slows down and is overtaken by \ac{amcl}.
The steady state performance is also comparable.
While the performance temporarily degrades, it manages to recover and keep a steady error rate throughout the whole run.
On the other hand, the harsher ghost factor \ensuremath{\acs{ghtfac} = 7.0} takes longer to converge, but remains steady and eventually outperforms the milder ghost factor.
Table \ref{sec:res:tbl:global} shows the \ac{rmse}, error along with other statistics.

\subsubsection{Qualitative Results}
Similar to the previous section, we can provide qualitative analysis by looking at the convergence behaviour and the estimated paths.

In order to visualise the convergence behaviour, figure \ref{sec:res:fig:conv_1} shows a series of time steps during the filters' initialisation. 
On the first image, the particles have been spread over the ground floor of a $(49\text{ x }49)\text{m}$ office area.
In this dataset, the robot is looking directly at a door during the beginning of the sequence.
Therefore, in figure \ref{sec:res:fig:conv_3} the filter converges with particles looking at doors that are a similar distance away.
The robot then proceeds to move through the doors.
Going through the door makes the filter converge significantly faster as it implicitly uses the ghost factor in the motion model.
It also gives the robot a more unique distribution of doors (on a corner), which makes the filter converge quickly.
This is shown in figure \ref{sec:res:fig:conv_4}, where the filter converges.

The estimated paths can be seen in figure \ref{sec:res:fig:path:gbl}, where the blue circle denotes the point of convergence.
It can be seen that AMCL takes longer to converge (further away from the corner room) than the range-based approach.
More importantly, it can be seen that the range-based approach suffers no noticeable degradation in the estimated trajectory over the room-level initialisation.
On the other hand, the ray-based method's performance degrades more noticeably.
This is because the filter converges in a long corridor with ambiguous label distributions (doors left and right are similarly spaced).
However, once the robot turns around the system recovers and performs comparably to the range-based approach.

As mentioned previously, entering or exiting rooms helps the filter converge because it can use the ghost factor in the motion model.
The following experiments, evaluate how the ghost factor affects the performance of the approach.

	\vspace{-0.2cm}
\subsection{Ghost Factor}\label{sec:res:ght}
	\vspace{-0.1cm}
The effect of the ghost factor can be measured in a similar way to the overall filter performance.
We show that the ghost factor provides more discriminative information when it is \textit{not} defined in a binary fashion.
This is shown in the label-only scenario for both the range-based and ray-based approaches, in both the global and room-level initialisation.
\subsubsection{Global Initialisation}
\begin{figure}[t]
  \begin{center} 
    \subfloat[Range-Based]{\includegraphics[width=0.48\linewidth]{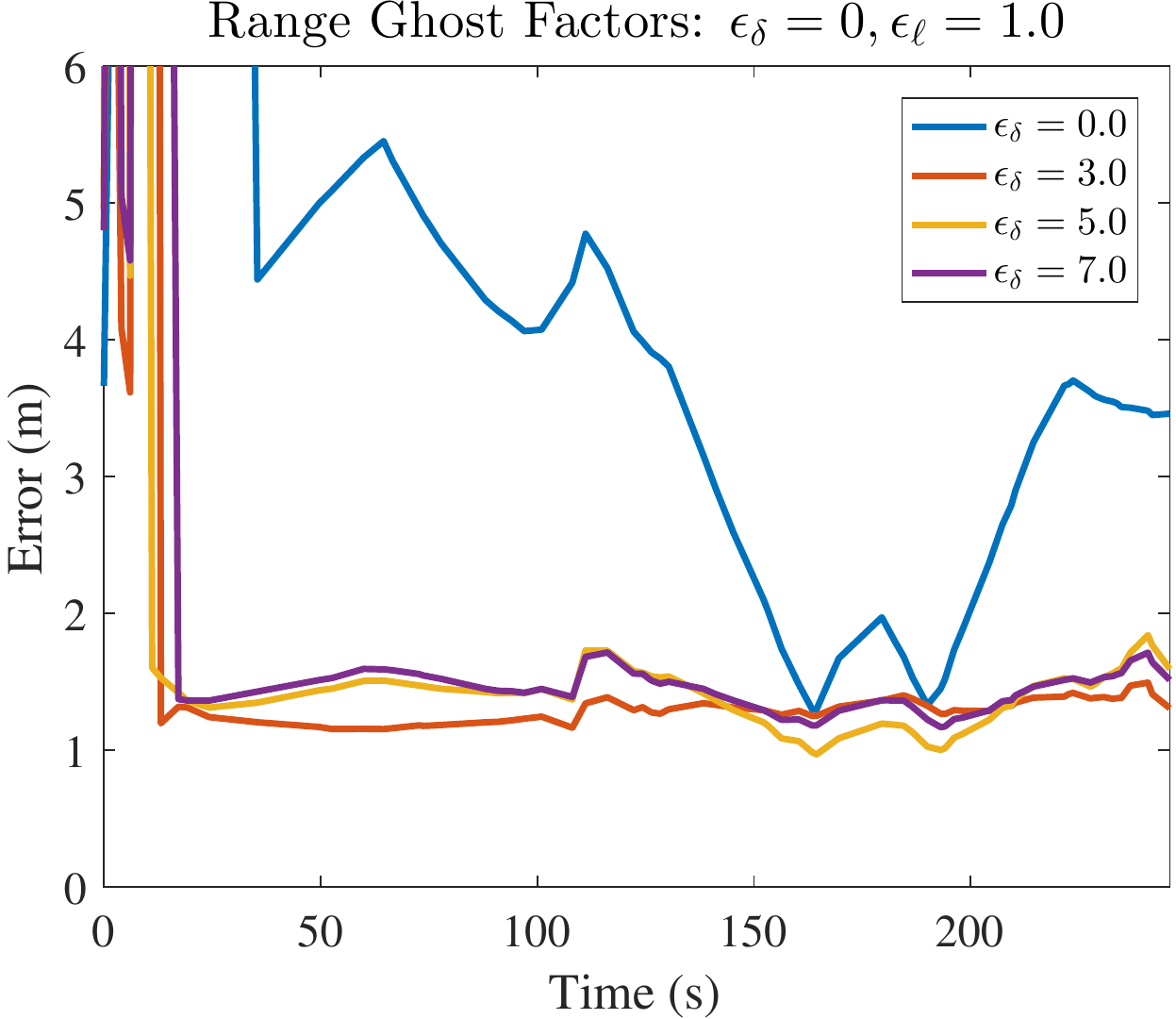}}\label{sec:res:fig:ghostfactor:range:global}
    \subfloat[Ray-Based]{\includegraphics[width=0.49\linewidth]{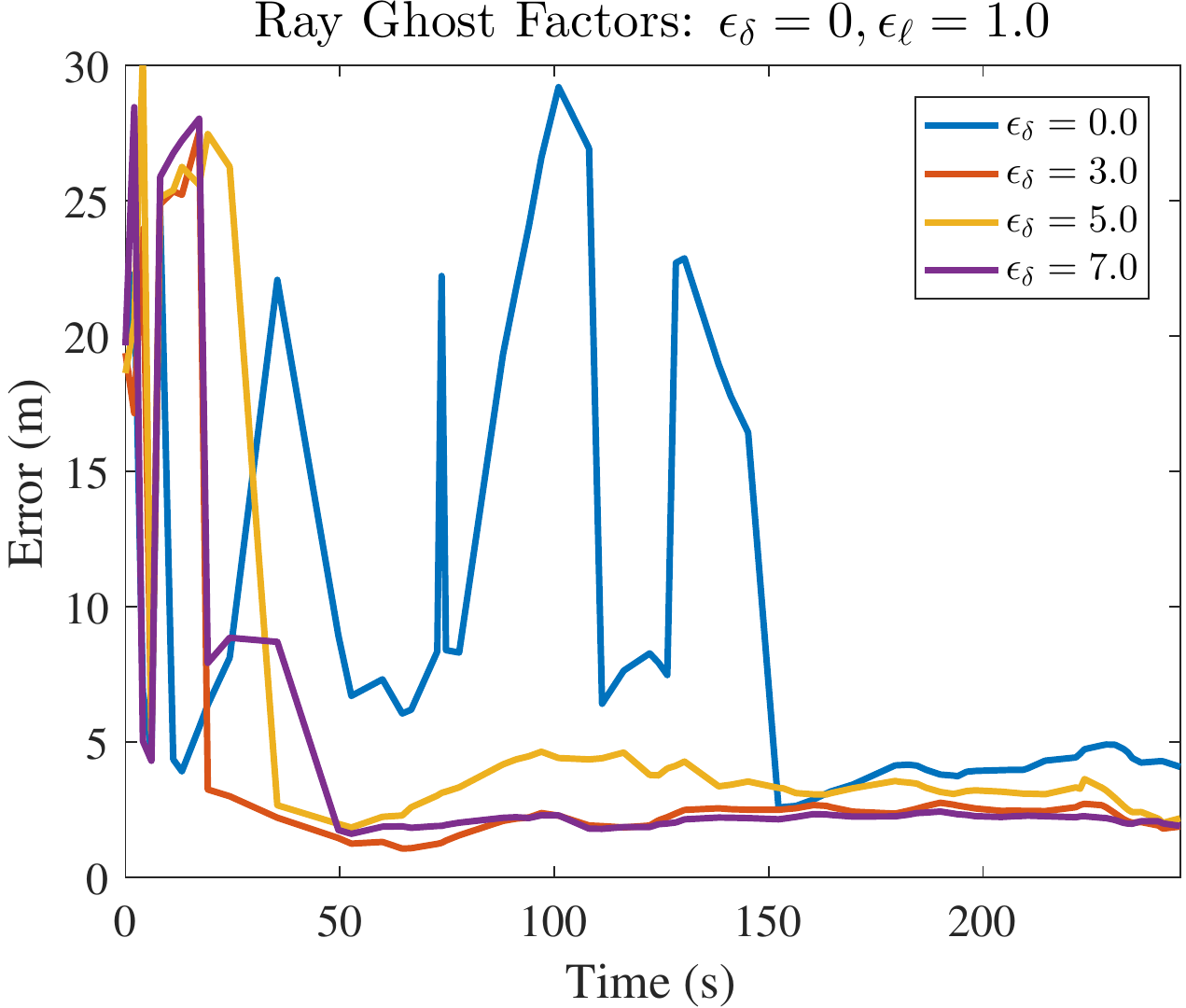}}\label{sec:res:fig:ghostfactor:ray:global}
  \end{center}
	\vspace{-0.3cm}
  \caption{Different ghost factors (\ensuremath{\acs{ghtfac}}), global initialisation.}
	\vspace{-0.4cm}
\label{sec:res:fig:ghostfactor:global}
\end{figure}

Figure \ref{sec:res:fig:ghostfactor:global} shows the effect of varying the ghost factor during global initialisation.
It can be seen that not penalising particles going through walls, (\ensuremath{\acs{ghtfac} = 0}), is not a good choice.
This makes sense, as there is very little to be gained from allowing particles to traverse occupied cells without any consequence.
It follows that  we should set the ghost factor as high as possible.
However, setting the ghost factor to a large value \ensuremath{(\acs{ghtfac} = 7.0)}, which corresponds to reducing the probability by \ensuremath{95\%} at \ensuremath{0.43\si{\meter}}, does not provide the best results.

While it might seem intuitive to assume that a higher \ac{ghtfac} will always be better, this is not the case.
High values of the ghost factor correspond to a binary interpretation of occupancy which makes \ac{mcl} systems unstable in the presence of discrepancies between the map and the environment.
This happens because otherwise correct particles can clip door edges and be completely eliminated from the system.
A harsh ghost factor also exacerbates problems with limited number of particles.
In fact, \ensuremath{\acs{ghtfac} = 3.0}, corresponding to a \ensuremath{95\%} reduction at \ensuremath{1.0\si{\meter}}, consistently showed the best results in all of the global initialisation experiments, as can be seen in table \ref{sec:res:tbl:ghostfactor:global}.

\subsubsection{Room-Level Initialisation}
\begin{figure}[t]
\vspace{-0.013\textheight}
  \begin{center}  
    \subfloat[Range-Based]{\includegraphics[width=0.49\linewidth]{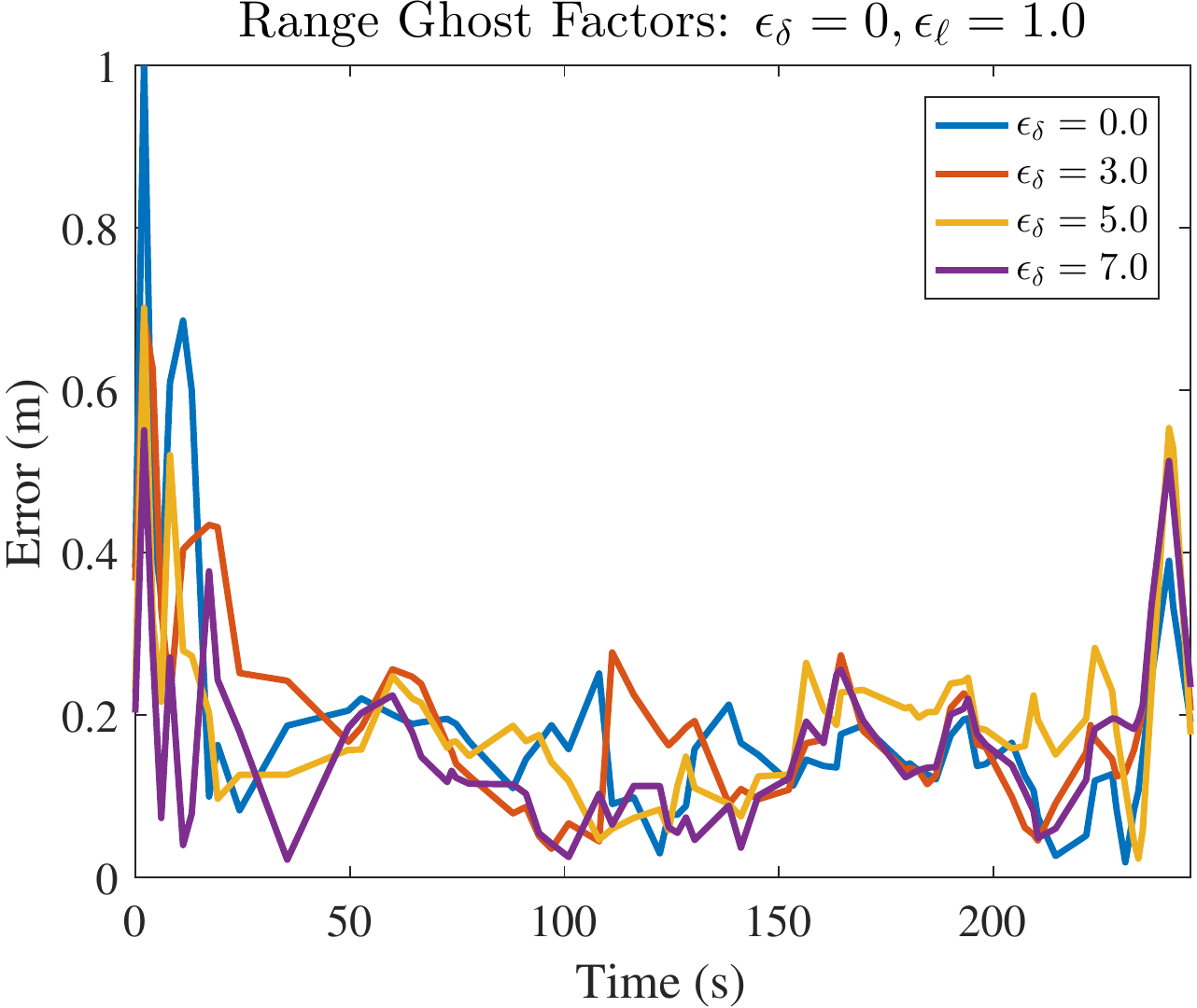}\label{sec:res:fig:ghostfactor:range:local}}
    \subfloat[Ray-Based]{\includegraphics[width=0.49\linewidth]{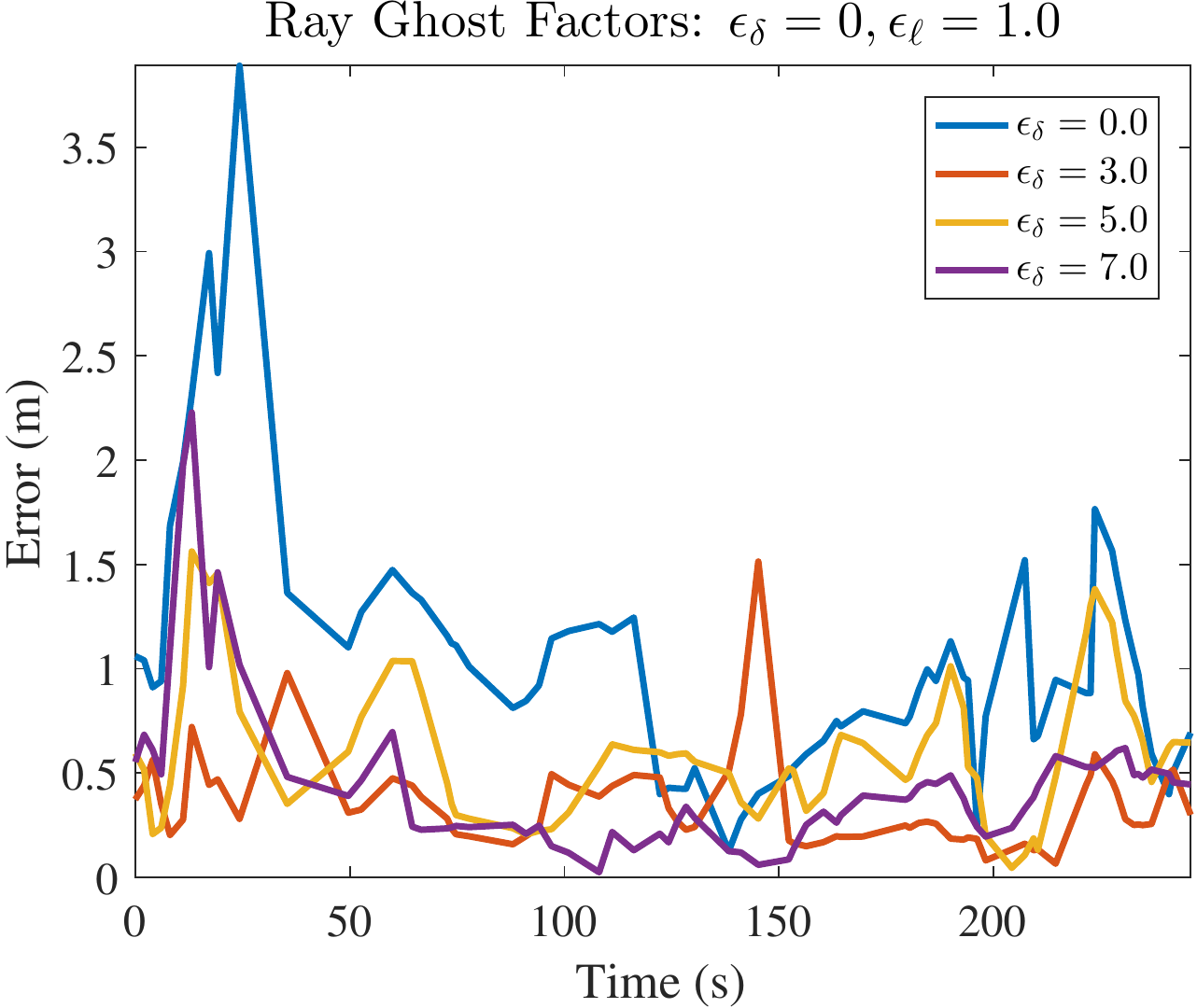}\label{sec:res:fig:ghostfactor:ray:local}}
  \end{center}
	\vspace{-0.3cm}
  \caption{Different ghost factors (\ensuremath{\acs{ghtfac}}), room-level initialisation.}
\label{sec:res:fig:ghostfactor:local}
	\vspace{-0.75cm}
\end{figure}

\begin{table}[tb]

\centering
\resizebox{\linewidth}{!}{
\begin{tabular}{|c|c|c|c|c|c|c|}
\hline
\multicolumn{7}{|c|}{\textbf{Average Trajectory Error (m)}}                                                                                       \\ \hline
\textbf{Approach}       		& \textbf{\acs{rmse}}	& \textbf{Mean}     	& \textbf{Median}   	& \textbf{Std. Dev.} 	& \textbf{Min}      	& \textbf{Max}       \\ \hline
AMCL                    		& 7.31          	& 2.26          	& \textbf{0.20} 	& 6.95           	& \textbf{0.028}	& 35.45          \\ \hline
\textbf{Range} (Label Only)      	& 6.71          	& 2.59          	& 1.31          	& 6.20           	& 1.15          	& 38.60          \\ \hline
\textbf{Range} (Combined)        	& \textbf{4.78} 	& \textbf{1.69} 	& 0.69          	& \textbf{4.47}  	& 0.43          	& \textbf{31.19} \\ \hline
\textbf{Rays} \ensuremath{(\acs{ghtfac} = 3.0)} 	& 7.74          	& 4.36          	& 2.46          	& 6.40           	& 1.07          	& 27.55          \\ \hline
\textbf{Rays} \ensuremath{(\acs{ghtfac} = 7.0)} 	& 8.09          	& 4.49          	& 2.22          	& 6.73           	& 1.61          	& 28.47          \\ \hline
\end{tabular}
}
	\vspace{-0.1cm}
\caption{Global Initialisation}
	\vspace{-0.1cm}
\label{sec:res:tbl:global}
\end{table}
\begin{table}
\centering
\resizebox{\linewidth}{!}{%
\begin{tabular}{|c|c|c|c|}
\hline
\multicolumn{4}{|c|}{\textbf{Average Trajectory Error (RMSE)}}                                                   \\ \hline
\textbf{Ghost Factor \ensuremath{(\acs{ghtfac})}} 	& \textbf{Range (Labels)} 	& \textbf{Range (Weighted)} 	& \textbf{Rays}     \\ \hline
0.0                                    	& 10.88                	& 10.13                 	& 11.71         \\ \hline
3.0                           		& \textbf{6.71}       	& \textbf{4.78}         	& \textbf{7.74} \\ \hline
5.0                                    	& 6.97                	& 6.30                  	& 9.54          \\ \hline
7.0                                    	& 7.19                	& 6.10                  	& 8.09          \\ \hline
\end{tabular}%
}
	\vspace{-0.1cm}
\caption{Global \ac{ate} for Different Ghost Factors}
	\vspace{-0.4cm}
\label{sec:res:tbl:ghostfactor:global}
\end{table}
	\vspace{-0.1cm}
In terms of room-level initialisation, having an aggressive ghost factor is more in line with our initial intuition.
Table \ref{sec:res:tbl:ghostfactor:local} shows that for both of the range-based scenarios, \ensuremath{\acs{ghtfac} = 7.0} provides the best results.
This is because room-level initialisation in the presence of range-based measurements is a much easier problem to solve.
On the other hand, the ray-based scenario still prefers a milder ghost factor of \ensuremath{\acs{ghtfac} = 3.0}.
In this scenario, inaccuracies in both the map and the sensing modalities allow for otherwise correct particles to be heavily penalised by an aggressive ghost factor.
Both of these results are reflected in figures \ref{sec:res:fig:ghostfactor:range:local} and \ref{sec:res:fig:ghostfactor:ray:local}.\looseness=-1

These results allow us to come to a single conclusion.
The ghost factor must be tuned to the expected amount of noise in the map and sensing modality.
Aggressive ghost factors can be used in cases where the pre-existing map is accurate and densely sampled, such as the case where the map was collected by the same sensor being used to localise (\ie \ac{slam}).
On the other hand, where there are expected differences between the map and what the robot observes (\eg furniture, scale errors, etc.), it is beneficial to provide a milder ghost factor and to be more lenient to small pose errors.\looseness=-1
\vspace{-0.25cm}
\subsection {Timing}
	\vspace{-0.1cm}
The speed of our approach was evaluated on a machine equipped with an Intel Xeon X5550 (2.67GHz) and an NVidia Titan X (Maxwell). 
During room-level initialisation, or once the system has converged, our approach can run with $250$ particles in $~10$ms, leaving us more than enough time to process the images from the Kinect into a \ac{sedar} scan.
Transforming the RGB images into semantic labels is the most expensive operation, taking on average $~120$ms. 
This means that a converged filter can run at $8-10$ fps. 
When performing global localisation, we can integrate a new sensor update, using $50,000$ particles, in $~2.25$ seconds. As \ac{mcl}-based approaches require motion between each sensor integration, this is still effectively near real-time, and orders of magnitude faster than competing vision approaches. 

	\vspace{-0.2cm}
\section{Conclusion}
	\vspace{-0.2cm}
In conclusion, this work has demonstrated that human-inspired localisation based on distinctive landmarks, is an effective alternative to traditional scan-matching. 
We demonstrated how the semantic information provided by \ac{sedar} could be utilised in both the motion model and the sensor model (with and without range data). 
Our experiments show that this new information is highly complementary to state-of-the-art techniques, providing a 35\% reduction in errors over either technique alone. 
Based on this compelling evidence, we can conclude the application of \ac{sedar} (and semantic information in general) should be explored further within the wider field of robotics.

More generally, this work reinforces the conclusions of other recent research: machine learning has now reached the point where the subjective aspects of biological perception (such as semantic scene understanding) can be reliably emulated. 
As such, the biologically-inspired paradigm which has long been a staple of robot hardware design, is now also feasible (and essential) for robot software design.

To this end an interesting avenue for future work would be to follow recent research in visual odometry, and utilise single-image depth and/or surface normal estimation techniques for localisation. 
This could implicitly detect scene elements of known sizes, which is another vital component of biological perception. %
\begin{table}[t]
\centering
\resizebox{\linewidth}{!}{%
\begin{tabular}{|c|c|c|c|}
\hline
\multicolumn{4}{|c|}{\textbf{Average Trajectory Error (RMSE)}}                                                   \\ \hline
\textbf{Ghost Factor \ensuremath{(\acs{ghtfac})}} 	& \textbf{Range (Labels)} 	& \textbf{Range (Weighted)} 	& \textbf{Rays}     \\ \hline
0.0                                    	& 0.25               		& 0.27                  	& 1.20          \\ \hline
3.0                                    	& 0.24                		& 0.25                  	& \textbf{0.40} \\ \hline
5.0                                    	& 0.22                		& 0.24                  	& 0.70          \\ \hline
7.0                                    	& \textbf{0.19}       		& \textbf{0.22}         	& 0.58          \\ \hline
\end{tabular}%
}
	\vspace{-0.1cm}
\caption{Room-Level \ac{ate} for Different Ghost Factors}
	\vspace{-0.8cm}
\label{sec:res:tbl:ghostfactor:local}
\end{table}
\vspace{-0.6cm}
{
\bibliographystyle{plain}
\bibliography{references}
}

\end{document}